\documentclass[manuscript,screen]{acmart}

\AtBeginDocument{%
  }

\usepackage{adjustbox} 
\usepackage{arydshln}
\usepackage{subcaption}
\usepackage{dashrule}
\usepackage{ragged2e}
\usepackage{booktabs}    
\usepackage{multirow}    
\usepackage{tabularx}    
\usepackage{makecell}    
\usepackage{color}
\usepackage{url}

\DeclareUnicodeCharacter{2010}{-} 
\DeclareUnicodeCharacter{2212}{-} 
\DeclareUnicodeCharacter{202F}{\,} 
\usepackage{geometry}
\newcolumntype{Y}{>{\centering\arraybackslash}X}

\begin{document}

\title{Knowledge Graphs Meet Graph Neural Networks: A Comprehensive Survey}



\author{Chengcheng Sun}
\orcid{0000-0002-0995-9985}
\affiliation{%
  \institution{School of Safety Engineering, China University of Mining and Technology}
  \city{Xuzhou}
  \state{Jiangsu}
  \country{China}
}
\email{scc@cumt.edu.cn}

\author{Jiayun Tian*}
\orcid{0009-0005-4716-9449}
\affiliation{%
  \institution{School of Computer Science and Technology, China University of Mining and Technology}
  \city{Xuzhou}
  \state{Jiangsu}
  \country{China}
}
\email{tianjy@cumt.edu.cn}

\author{Cheng Zhai*}
\orcid{0000-0001-9256-1494}
\affiliation{%
  \institution{State Key Laboratory of Coal Mine Disaster Prevention and Control, China University of Mining and Technology}
  \city{Xuzhou}
  \state{Jiangsu}
  \country{China}
}
\email{greatzc@cumt.edu.cn}

\author{Zhixiao Wang*}
\orcid{0000-0002-4256-1477}
\affiliation{%
  \institution{School of Computer Science and Technology, China University of Mining and Technology}
  \city{Xuzhou}
  \state{Jiangsu}
  \country{China}
}
\email{zhxwang@cumt.edu.cn}

\author{Yajie Song}
\orcid{0009-0009-6265-9939}
\affiliation{%
  \institution{School of Computer Science and Technology, China University of Mining and Technology}
  \city{Xuzhou}
  \state{Jiangsu}
  \country{China}
}
\email{songyajie@cumt.edu.cn}

\author{Xiaobin Rui}
\orcid{0000-0003-0951-1512}
\affiliation{%
  \institution{School of Computer Science and Technology, China University of Mining and Technology}
  \city{Xuzhou}
  \state{Jiangsu}
  \country{China}
}
\email{ruixiaobin@cumt.edu.cn}

\author{Jian Zhang}
\orcid{0000-0003-4995-6495}
\affiliation{%
  \institution{School of Computer Science and Technology, China University of Mining and Technology}
  \city{Xuzhou}
  \state{Jiangsu}
  \country{China}
}
\email{zhangjian10231209@cumt.edu.cn}

\author{Philip S. Yu}
\orcid{0000-0002-3491-5968}
\affiliation{%
  \institution{Department of Computer Science, University of Illinois at Chicago}
  \city{Chicago}
  \state{IL}
  \country{US}
}
\email{psyu@uic.edu}

\renewcommand{\shortauthors}{C. Sun et al.}

\begin{abstract}
Abstract: Graph Neural Networks (GNNs) have emerged as a powerful paradigm in Knowledge Graphs (KGs) due to their intrinsic ability to model graph-structured data. However, there remains a lack of a systematic review about GNN-based methodologies across the entire knowledge graph technologies pipeline. To address this gap, we first propose a novel two-level taxonomy framework for GNN-based knowledge graph technologies: the KG technologies pipeline and GNN-based perspective. Specifically, the knowledge graph technologies pipeline covers knowledge graph construction, knowledge graph embedding, knowledge reasoning and knowledge graph applications. Meanwhile, the GNN-based perspective provides a new categorization of knowledge graph technologies with GNN models, such as GCN, GAT, and HGNN. Then, we analyze the advantages of GNN technology based on the characteristics of different tasks in the knowledge graph lifecycle. Furthermore, we detailed review various GNN-based models for knowledge graph following the proposed taxonomy, and summarize strengths and limitations. Finally, we discuss unresolved challenges and outline promising directions for future research.
\end{abstract}


\ccsdesc[500]{Computing methodologies~Artificial intelligence~Knowledge representation and reasoning}

\keywords{Knowledge Graphs, Graph Neural Networks, Knowledge Graph Construction, Knowledge Graph Embedding, Knowledge Reasoning}


\maketitle

\section{Introduction}
\label{introduction}
Knowledge serves as the foundation for human cognition and problem solving, embodying the accumulation of information and the transmission of wisdom. The study of knowledge representation and reasoning aims to provide intelligent systems with methods to express knowledge, enabling them to tackle complex tasks effectively \cite{peng2023knowledge}. As an essential practice in knowledge engineering, knowledge graphs construct semantic knowledge networks, addressing the limitations of traditional knowledge storage and utilization. They also act as a vital bridge between information and intelligence. Knowledge graphs are not only of importance in theoretical research, but also demonstrate tremendous potential in practical applications. They serve as a key driving force in advancing artificial intelligence and the development of the knowledge-based society. 

\begin{figure}[!htbp]
\centering
\includegraphics[width=0.85\linewidth]{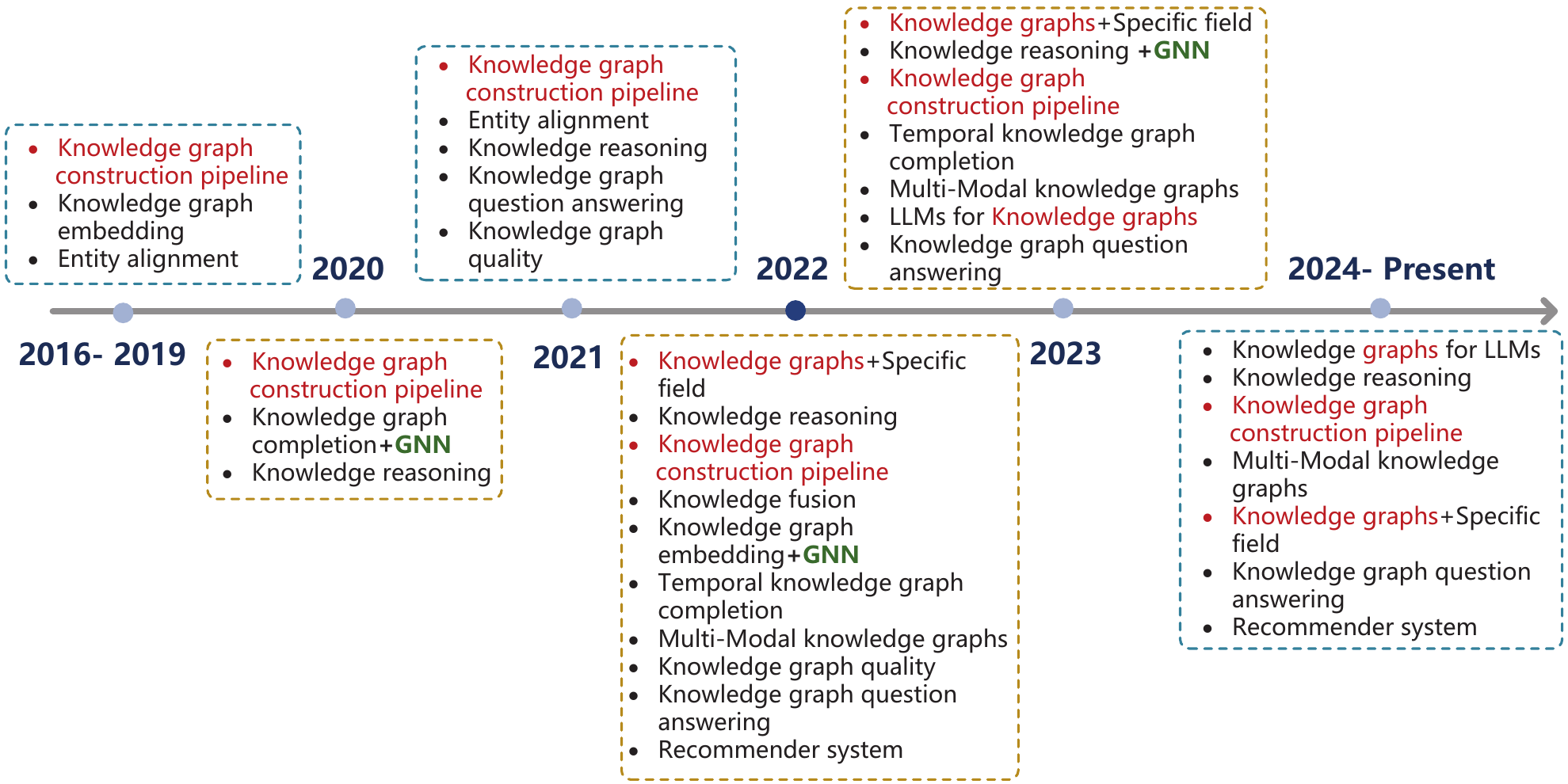}
\Description{Timeline of the knowledge graph review development.}
\caption{Timeline of the knowledge graph review development.}
\label{fig:Timeline}
\end{figure}

As an important form of knowledge representation, knowledge graphs have become a prominent research trend in academia and industry\cite{kong2022bolt}. 
As illustrated in Figure \ref{fig:Timeline}, we have organized the development trajectory of knowledge graph review studies from 2016 to present. 
Since the introduction of knowledge graphs, research reviews focusing on the knowledge graph construction process have consistently been a key area of interest in the academic community \cite{wu2018survey, zhao2024survey, abu2021domain, DBLP:conf/coling/TurkiOTBA24, ye-etal-2022-generative, zhu2022multi}. Hogan et al. \cite{hogan2021knowledge}, Ji et al. \cite{ji2021survey}, and Zhong et al. \cite{zhong2023comprehensive} systematically summarize the latest research advancements in knowledge graph construction from different perspectives and technical aspects. With the deepening of research, the related fields of knowledge graph are expanding. The topics such as knowledge completion \cite{DBLP:journals/pami/LiangMLLTWZLSH24}, knowledge reasoning \cite{shen2022comprehensive}, and knowledge representation \cite{cao2024knowledge} are gradually becoming important research directions. In recent years, the application of knowledge graphs has made significant progress, particularly in the context of the integration of artificial intelligence and machine learning. Knowledge graphs have become increasingly widespread in applications such as intelligent question answering \cite{khan2022categorization,   DBLP:journals/semweb/PerevalovBN24}, personalized recommendation systems\cite{guo2020survey, DBLP:conf/icde/GuoZQZ00023}, and natural language understanding \cite{wang2023fusing, DBLP:journals/tkde/HuLZHNL24, DBLP:conf/ijcnlp/SchneiderSVGSM22}. Furthermore, they have demonstrated vital value in industries such as smart healthcare\cite{abu2023healthcare, zhong2023knowledge}
, education \cite{fettach2022knowledge}, and news \cite{opdahl2022semantic, DBLP:conf/cikm/LiuBLZSWX19}.

However, the development of knowledge graphs also faces several technical challenges, such as scalability and knowledge incompleteness \cite{DBLP:conf/ijcai/Cai0GZLL23, DBLP:journals/tkde/XueZ23}. Knowledge graphs are essentially graph structures, where nodes represent entities and edges denote the relationships between these entities \cite{cheng2022financial}. GNNs demonstrate remarkable potential in facing these challenges in knowledge graphs due to their intrinsic ability to model graph-structured data. By propagating and aggregating neighborhood information through message-passing mechanisms, GNNs can effectively capture high-order topological patterns of nodes and edges\cite{khemani2024review}, thereby enhancing the latent representation of both entities and relations in multidimensional embedding spaces. Particularly, GNNs can automatically uncover latent patterns and regularities from vast amounts of graph-structured data, optimizing key tasks such as entity alignment \cite{zhang2021comprehensive} and relation extraction \cite{liu2022survey}. For  application, GNN architectures enable advanced semantic reasoning through their capacity to preserve graph context. Recent studies demonstrate their effectiveness in knowledge reasoning \cite{chen2020review}, knowledge graph question answering(KGQA) \cite{DBLP:journals/semweb/PerevalovBN24}, and recommendation systems \cite{gao2020deep}. 
Therefore, a thorough analysis of the benefits of combining GNNs with knowledge graphs is crucial for advancing the development and widespread adoption of knowledge graph technologies . Some scholars have also investigated the integration of GNNs with knowledge graphs \cite{arora2020survey, chetoui2022graph, ma2024review}. They point out that the advantages of GNNs in deep learning help improve the data representation and reasoning capabilities of knowledge graphs. However, these studies focus on the integration of specific tasks within knowledge graphs and GNNs. There remains a lack of a systematic review about GNN-based methodologies across the entire knowledge graph technologies pipeline.

To the best of our knowledge, previous review papers\cite{shu2023entity,li2024graph,liang2024survey,guan2022event} have primarily focused on general knowledge graph issues or GNN techniques. However, they have not provided a comprehensive and systematic analysis of GNN-based methodologies across the entire knowledge graph technologies pipeline. To gain a deeper understanding of the role and potential of GNN in knowledge graph technologies, this paper conducts a novel two-level taxonomy framework: the KG technologies pipeline and and the GNN perspective. On the one hand, we discuss knowledge graph technologies throughout the knowledge graph lifecycle: knowledge graph construction, knowledge representation, knowledge reasoning, and  applications. On the other hand, we systematically investigate a wide range of the state-of-the-art graph learning methods based on GNNs, focusing on their deployment across the aforementioned KG tasks. More specifically, we investigate how researchers tailor graph structures and GNN architectures across tasks with diverse semantic characteristics to effectively capture the intrinsic semantic patterns within the knowledge, such as Graph Neural Network (GCN), Graph Attention Network (GAT), and Hypergraph Graph Neural Network (HGNN). Finally, we discuss unresolved challenges and outline promising directions for future research. The main contributions of this paper are as follows:
\begin{itemize}
\item \textbf{Graph neural network perspective:} A review of knowledge graphs is presented from the novel perspective of GNNs, highlighting recent advancements and methodologies in integrating GNNs with KGs technologies.
\item \textbf{Novel two-level taxonomy:} This paper proposes a novel two-level taxonomy framework for knowledge graph:  the KG technologies pipeline and the GNN perspective.
\item \textbf{Comprehensive review:} This paper provides a comprehensive review of knowledge graph technologies throughout the knowledge graph lifecycle, including knowledge graph construction, knowledge graph embedding, knowledge reasoning and knowledge graph applications.
\item \textbf{Open issue and future directions:} This paper also presents open issues and challenges of knowledge graph technologies in terms of data, methods and applications, which provide insights for advancing future research directions in this area.
\end{itemize}

The remainder of this survey is organized as follows. Section \ref{preliminaries} gives the preliminaries and notations of knowledge graphs and GNNs. Section \ref{taxonomy} provides a two-level taxonomy framework from the knowledge graphs construction pipeline and the perspective of GNNs. Section \ref{KGC} presents a construction technical overview of knowledge graphs. Sections \ref{KGE}, \ref{KR}, and \ref{KGA} provide an overview of knowledge representation, knowledge reasoning, and the applications of knowledge graphs, respectively. We suggest promising future research directions in Section \ref{issues}, while Section \ref{conclusion} comes to the conclusion of this paper. 

\section{Preliminaries and notations}
\label{preliminaries}

\subsection{Knowledge Graph and Technologies Pipeline}
Knowledge graph is a knowledge representation method based on graph structure\cite{singhal2012introducing}, where entities (e.g. people, places, events, etc.) and their relationships are organised and represented by means of nodes and edges \cite{bordes2011learning}. 
Each node represents an entity and edges represent various semantic relationships between entities. 
A knowledge graph can be formally defined in the form of triples, i.e., $G = (E1, R, E2)$, where $E1$ is the head entity, $E2$ is the tail entity, and $R$ represents the relationship between the two entities.

\begin{figure}[!htbp]
\centering
\includegraphics[width=1\linewidth]{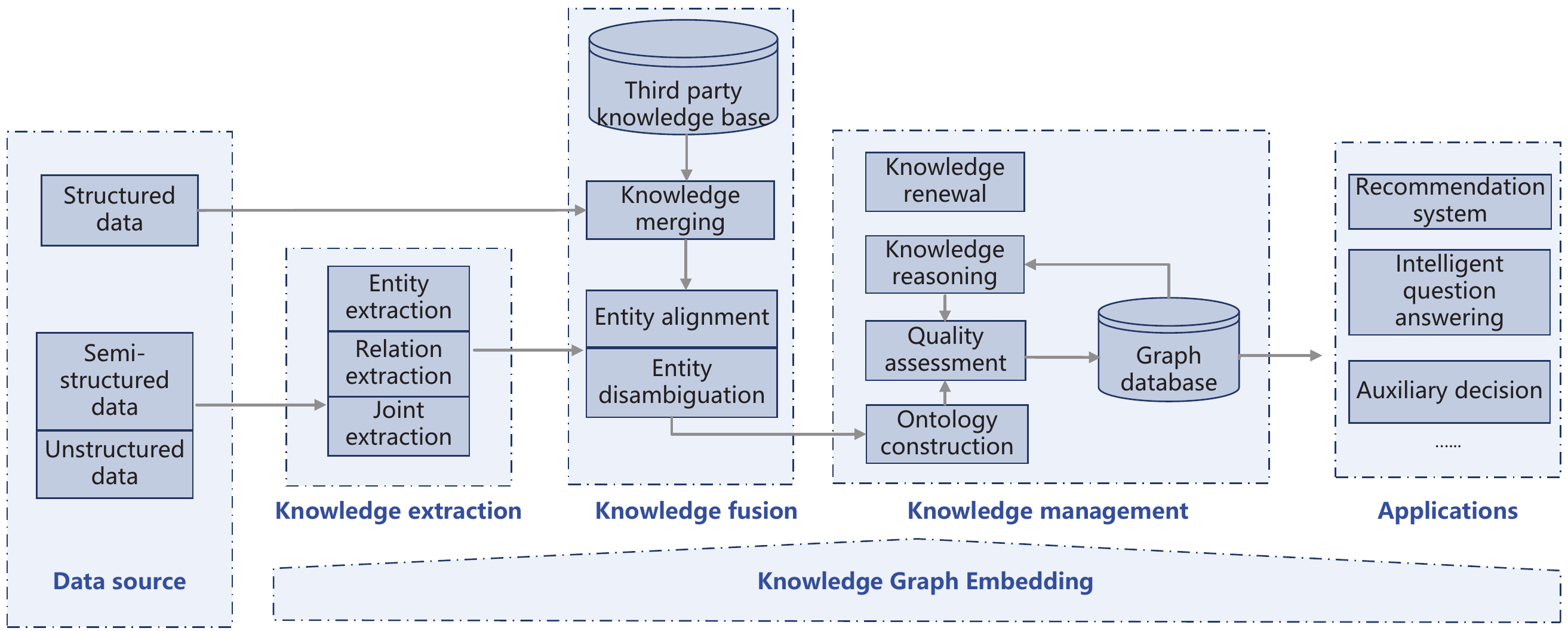}
\Description{Knowledge graph technologies pipeline.}
\caption{Knowledge graph technologies pipeline.}
\label{fig:Construction Process}
\end{figure}

At present, a large number of knowledge graphs have emerged, of which the representative ones are OpenKN \cite{jia2014openkn}, CN-Dbpedia \cite{xu2017cn}, DBpedia \cite{auer2007dbpedia}, Freebase \cite{bollacker2008freebase}, NELL \cite{carlson2010toward}, YAGO \cite{mahdisoltani2013yago3} and so on. In addition to the widespread development of open-domain knowledge graphs, many domain-specific knowledge graphs \cite{de2023harmonic, stranisci2023world, garcia2023holocaust, liu2024kg, xu2023seqcare} have also been proposed and gradually constructed in recent years. 
Figure \ref{fig:Construction Process} illustrates the pipeline of knowledge graph technologies. The workflow begins with knowledge extraction from data sources through entity extraction, relation extraction, and joint extraction. Subsequently, knowledge from diverse sources is integrated via knowledge merging, entity alignment, and disambiguation. The knowledge management phase follows, encompassing knowledge updating, reasoning, quality evaluation, and ontology construction to ensure accuracy and timeliness. The processed knowledge is then stored as triples in graph databases, supporting downstream applications such as recommendation systems and intelligent question answering. It is worth noting that knowledge graph embedding plays a pivotal role across various steps of knowledge graph technologies.

\subsection{Graph Neural Networks}
 
\begin{figure}[!htbp]
    \centering
    \begin{subfigure}{0.48\textwidth}
        \centering        \includegraphics[width=0.95\linewidth]{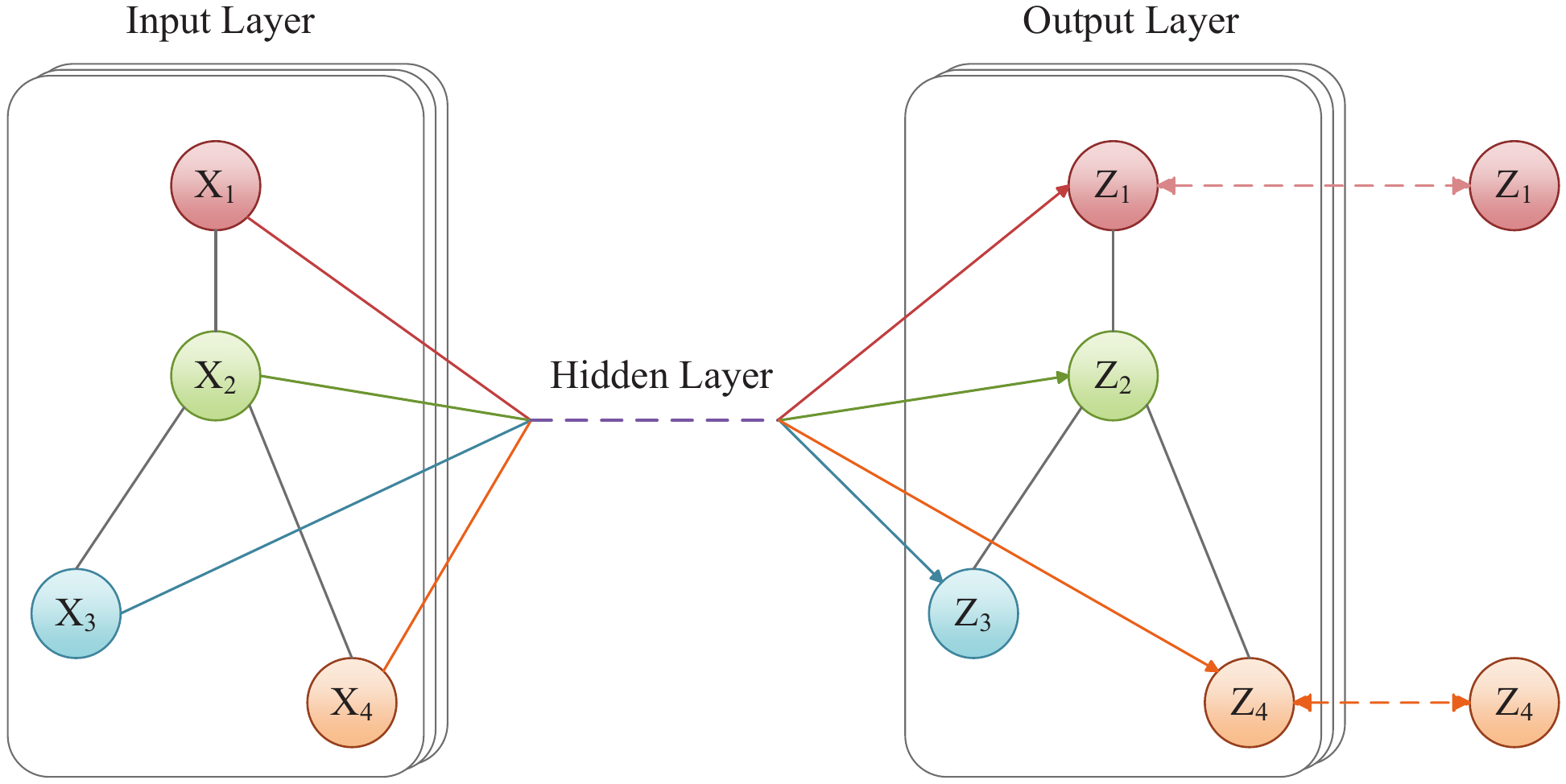}
        \caption{Basic structural form of GCN.}
        \label{fig:GCNa}
    \end{subfigure}
    \begin{subfigure}{0.48\textwidth}
        \centering        \includegraphics[width=0.95\linewidth]{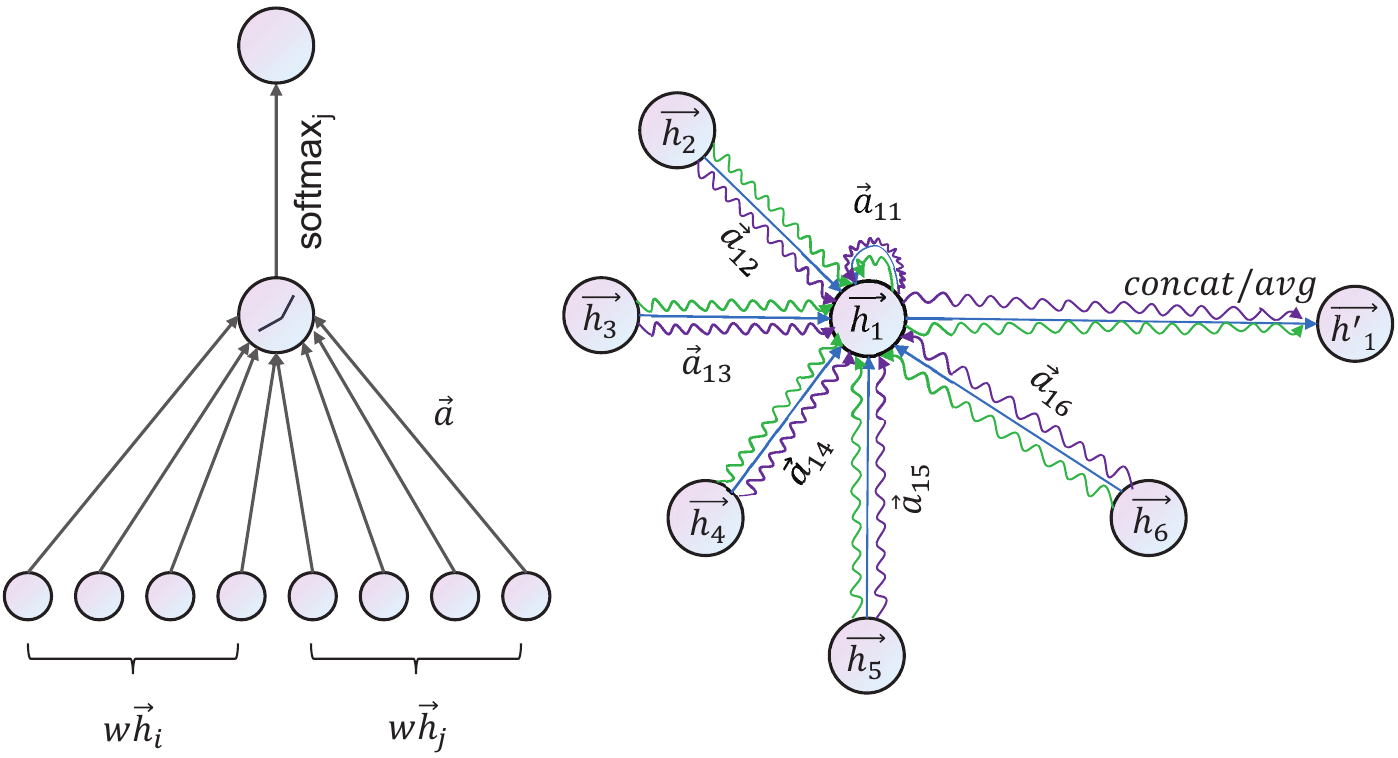}
        \caption{The attention mechanism in GAT.}
        \label{fig:GATb}
    \end{subfigure}    
    \Description{GCN and GAT.}
    \caption{GCN and GAT.}
    \label{fig:GCN_GAT}
\end{figure}

The core idea of GNNs is to propagate and learn information through the interactions between nodes and their neighborhood relationships \cite{gori2005new, zhang2020deep}. Depending on the aggregation mechanism, GNNs can be categorized into graph convolutional networks (GCN) and graph attention networks (GAT)\cite{wu2020comprehensive}. 
GCN was first introduced by Kipf and Welling in 2017, as shown in Figure \ref{fig:GCNa}. They applied the framework of spectral graph theory to propose an efficient method for propagating information via the adjacency matrix. 
In GCN, the representation of a node is obtained by performing a weighted sum of its neighboring nodes' features, followed by a nonlinear transformation \cite{bhatti2023deep}. Specifically, the update rule for GCN is as follows:
\begin{equation}H^{(l+1)} = \sigma \left( \hat{A} H^{(l)} W^{(l)} \right)\end{equation}
where, $H^{(l)}$ is the matrix of node representations at layer $l$. $\hat{A}$ is the normalized adjacency matrix, given by $D^{-1/2} (A + I) D^{-1/2}$, where $A$ is the adjacency matrix, $I$ is the identity matrix, and $D$ is the degree matrix. $W^{(l)}$ is the learnable weight matrix at layer $l$, and $\sigma$ is a nonlinear activation function, typically ReLU. Alternatively, the above equation can be rewritten from the node perspective as follows:
\begin{equation} h_{v}^{(l+1)} = \sigma\left(W^{(l)} \cdot \sum_{u \in \mathcal{N}(v)} h_{u}^{(l)} + b^{(l)}\right) \end{equation}
where, $h_{v}^{(l)}$ is the feature representation of node $v$ at the $l-th$ layer. ${N}(v)$ denotes the set of neighboring nodes of $v$. $W^{(l)}$ represents the weight matrix at the $l-th$ layer, and $b^{(l)}$ is the bias.

Unlike the GCN uniform weighting operation, GAT \cite{velickovic2017graph} captures local information in the graph structure by introducing a self-attention mechanism in Figure \ref{fig:GATb}. This allows GAT to assign different weights to the connections between a node and its neighbors based on the relationships and feature similarities between the nodes \cite{10.1007/s10462-023-10577-2}. 
The core idea of GAT is to compute attention coefficients for each neighboring node during information propagation. These coefficients are then used to weight and aggregate the features of neighboring nodes. The update rule for GAT is as follows:
\begin{equation} e_{ij} = LeakyReLU\left(a^{\top} \left[ W h_i \, \| \, W h_j \right] \right) \end{equation}
\begin{equation} \alpha_{ij} = \frac{\exp(e_{ij})}{\sum_{k \in \mathcal{N}(i)} \exp(e_{ik})} \end{equation}
\begin{equation} h_i^{(l+1)} = \sigma \left( \sum_{j \in \mathcal{N}(i)} \alpha_{ij} W^{(l)} h_j \right) \end{equation}
where, $e_{ij}$ is the attention score between node $i$ and its neighbor $j$, indicating their relevance. $a$ is the learnable attention weight vector. $W$ is the weight matrix applied to the node features. $\alpha_{ij}$ is the attention coefficient, representing the strength of the relationship between nodes $i$ and $j$. $\mathcal{N}(i)$ denotes the set of neighbors of node $i$. By assigning different attention weights to each neighbor, GAT can adaptively adjust the influence of neighboring nodes in the aggregation process\cite{2024LGAT}. This makes it particularly suitable for graphs with varying node relationships and complex structures.

\section{Taxonomy of GNN-based Knowledge Graph Techniques}
\label{taxonomy}

\begin{figure}[!htbp]
    \centering
    \includegraphics[width=1.0\linewidth]{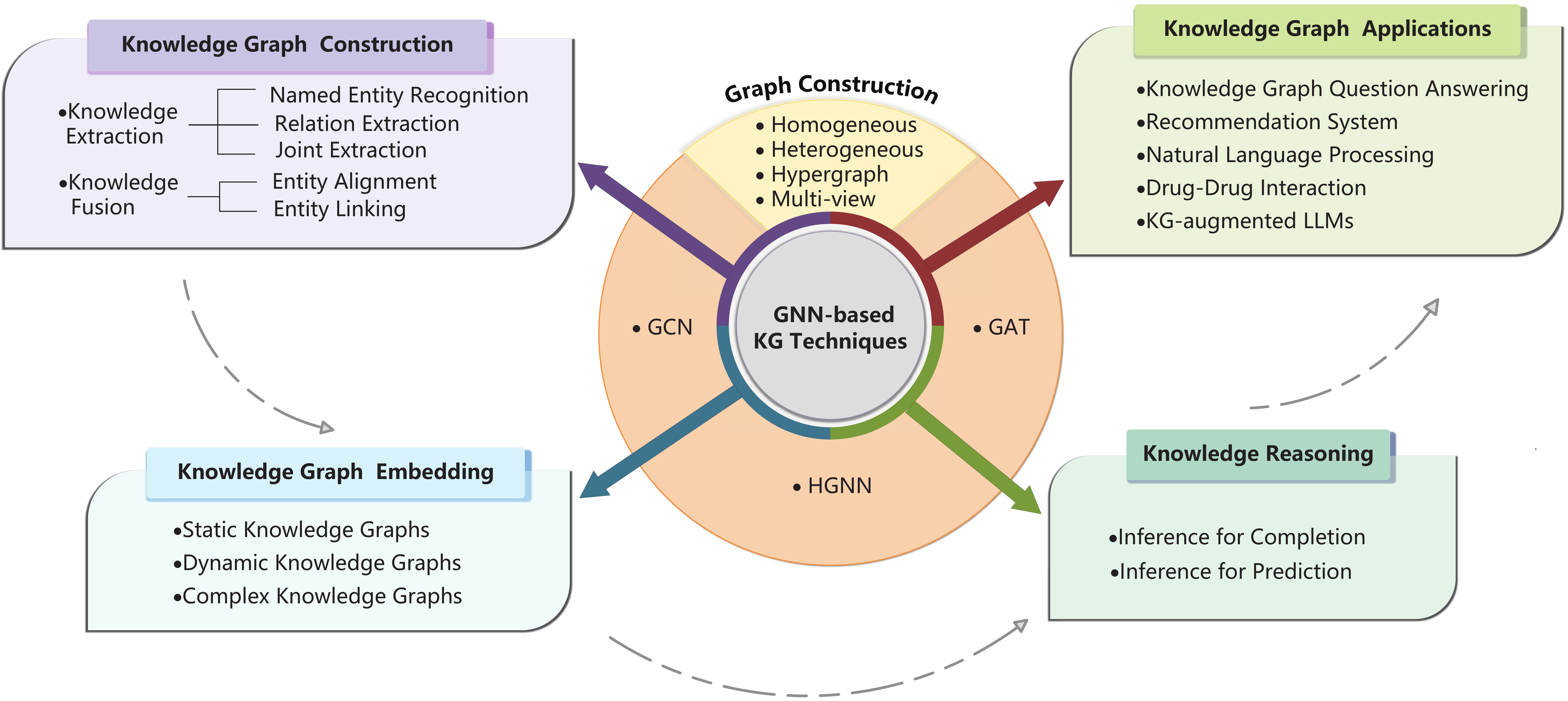}
    \Description{Taxonomy of GNN-based knowledge graph techniques.}
    \caption{Taxonomy of GNN-based knowledge graph techniques.}
    \label{fig:Taxonomy}
\end{figure}

In this section, we provide an overview of a novel two-level taxonomy framework: the KG technologies pipeline and GNN-based perspective. As shown in the figure \ref{fig:Taxonomy}, we categorize mainstream GNN architecture systems into representative frameworks such as GCN, GAT, and HGNN. Graph construction strategies encompass multiple paradigms ranging from fundamental homogeneous and heterogeneous graphs to advanced hypergraphs and multi-view graphs. Knowledge graphs techniques can be divided into four key steps: knowledge graph construction, knowledge representation, knowledge reasoning, and knowledge graph applications. At each step, we highlight the major tasks and corresponding typical GNNs methods. 

\textbf{Knowledge Graph Construction : }The core objective of the knowledge graph construction is to extract knowledge from diverse data sources and integrate this information to build the initial structure of the knowledge graph. Based on the task objectives, this step is divided into two main categories: knowledge extraction and knowledge fusion. 1) Knowledge extraction includes tasks such as named entity recognition \cite{xu2021supervised}, relation extraction \cite{DBLP:conf/emnlp/DongPL21}, and joint extraction \cite{yu2020joint}. These tasks aim to identify entities and their relationships from data, serving as the foundational step in knowledge graph construction. 2) Knowledge fusion focuses on entity alignment \cite{jiang2024toward} and entity linking \cite{zaporojets2022tempel}. It ensures that duplicate or ambiguous entities from different sources are unified and correctly linked, which is a critical step for knowledge integration. By leveraging graph representations and attention mechanisms, GNN-based methodologies significantly enhance task performance.

\textbf{Knowledge Graph Embedding : }The knowledge graph embedding aims to map the entities and relations in a knowledge graph into a low-dimensional vector space. This facilitates downstream tasks by providing a structured representation of the graph. Based on the type of knowledge graph and its application scenarios, this step is further divided into three categories. 1) Static Knowledge Graph \cite{jiang2022multiview}: This type of knowledge graph is suited for scenarios where both nodes and relationships are fixed. 
2) Dynamic Knowledge Graph \cite{gracious2021neural}: This type adapts to scenarios where nodes or edges change over time or according to different environments. 
3) Complex knowledge graph: Extend the traditional knowledge graph representation by integrating multiple types of relations \cite{DBLP:journals/corr/abs-2412-12158}, additional information \cite{DBLP:conf/emnlp/GalkinTMUL20}, and multimodal data \cite{zhang2023aspectmmkg} to build a richer graph representation. By capturing both local and global graph features, GNNs effectively enhance the performance of representation learning.

\textbf{Knowledge Reasoning : }The goal of the knowledge reasoning is to perform logical reasoning using the existing information in the knowledge graph, addressing tasks such as knowledge graph completion or new knowledge prediction. Based on the task objectives, this phase is divided into two main categories. 1) Inference for Completion: Infer missing triples in an existing knowledge graph by reasoning over known triples and predicting the missing ones \cite{zong2024renn}. 2) Inference for Prediction: Learn the unidirectional evolution patterns in the historical states of a temporal knowledge graph to effectively predict future states \cite{huang2024confidence}. GNNs are well-suited for this phase, as they can model the complex dependencies within the graph.

\textbf{Knowledge Graph Applications : }The application phase of knowledge graphs focuses on leveraging knowledge graph techniques to solve real-world problems. Based on the application objectives, this phase covers the following key areas: 1) Knowledge Graph Question Answering \cite{chen2023multi},
2) Recommendation Systems \cite{wang2024unleashing},
3) Natural Language Processing \cite{shi2023hallucination},
4) Drug-Drug Interaction \cite{lin2020kgnn}. 
5) KG-augmented LLMs. Due to its efficient modeling of graph-structured data, GNNs have become a key technology in knowledge graph applications, significantly improving task performance in complex scenarios.

\section{Knowledge Graph Construction}
\label{KGC}

\subsection{Knowledge Extraction}
Knowledge extraction primarily consists of Named Entity Recognition (NER) \cite{DBLP:conf/aaai/FuTCHH21} and Relation Extraction (RE) \cite{DBLP:conf/emnlp/HuZYLLWY21}. In recent years, joint extraction \cite{DBLP:conf/emnlp/YanZFZW21} has been proposed to simultaneously identify entities and their relationships. Below (a), (b) and (c) are the diagrams of NER, RE and joint extraction respectively. For each category, we classify existing research into several distinct branches, such as integrating external knowledge or addressing complex task scenarios, to reflect the diverse methodological landscape in the field. As summarized in Table \ref{tab:Knowledge Extraction}, we analyzed and refined these representative methods based on two core dimensions: composition strategies and model architecture.

\begin{figure}[!htbp]
    \centering
    \begin{minipage}[c]{0.47\linewidth}
        \centering
        \begin{subfigure}[b]{\linewidth}
            \centering            \includegraphics[width=\linewidth,height=8cm]{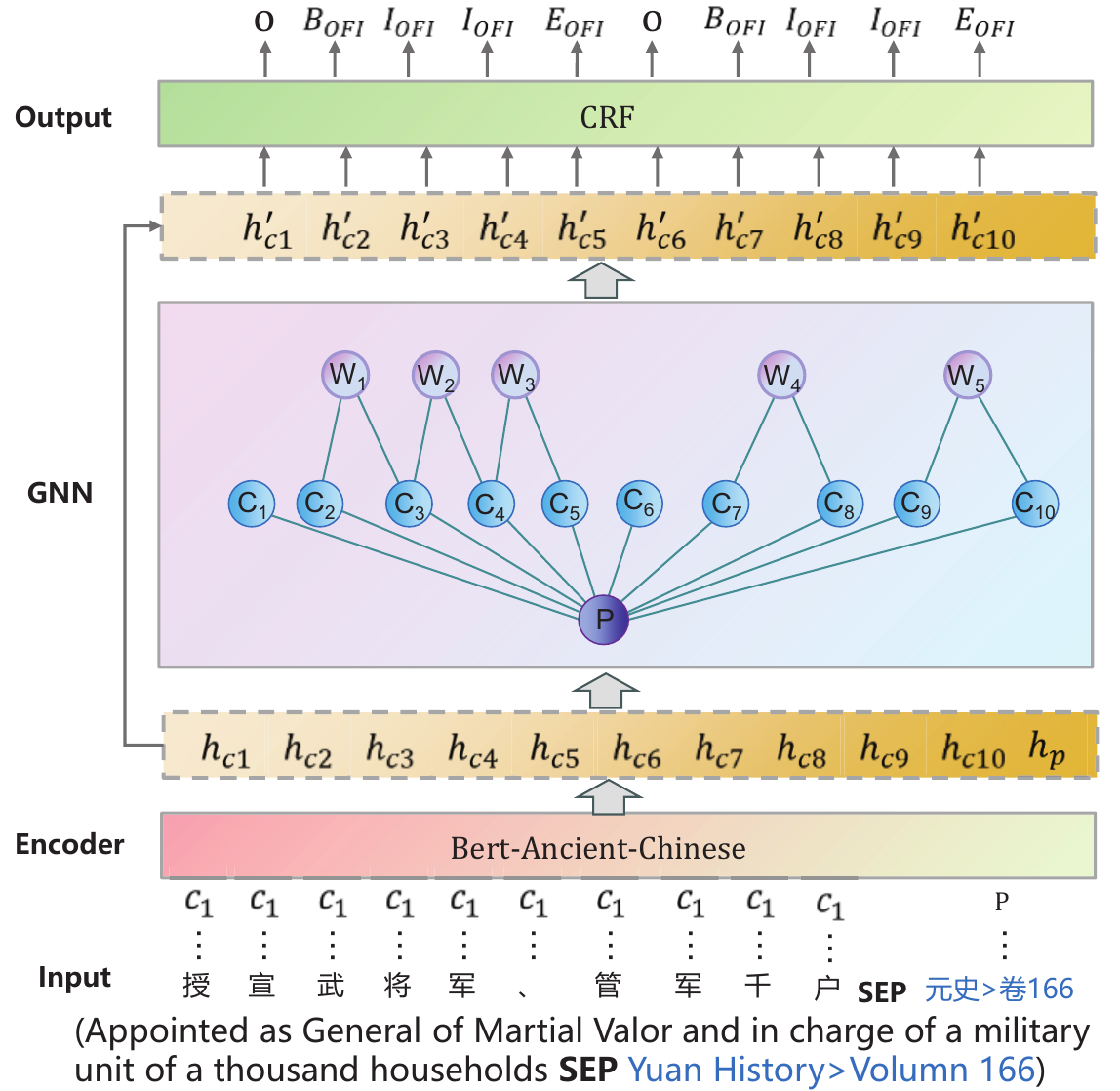}
            \caption{Named Entity Recognition.}
            \label{fig:NER_a}
        \end{subfigure}
    \end{minipage}
    \hfill 
    \begin{minipage}[c]{0.47\linewidth}
        \vspace{0pt} 
        \begin{subfigure}[b]{\linewidth}            \includegraphics[width=0.85\linewidth]{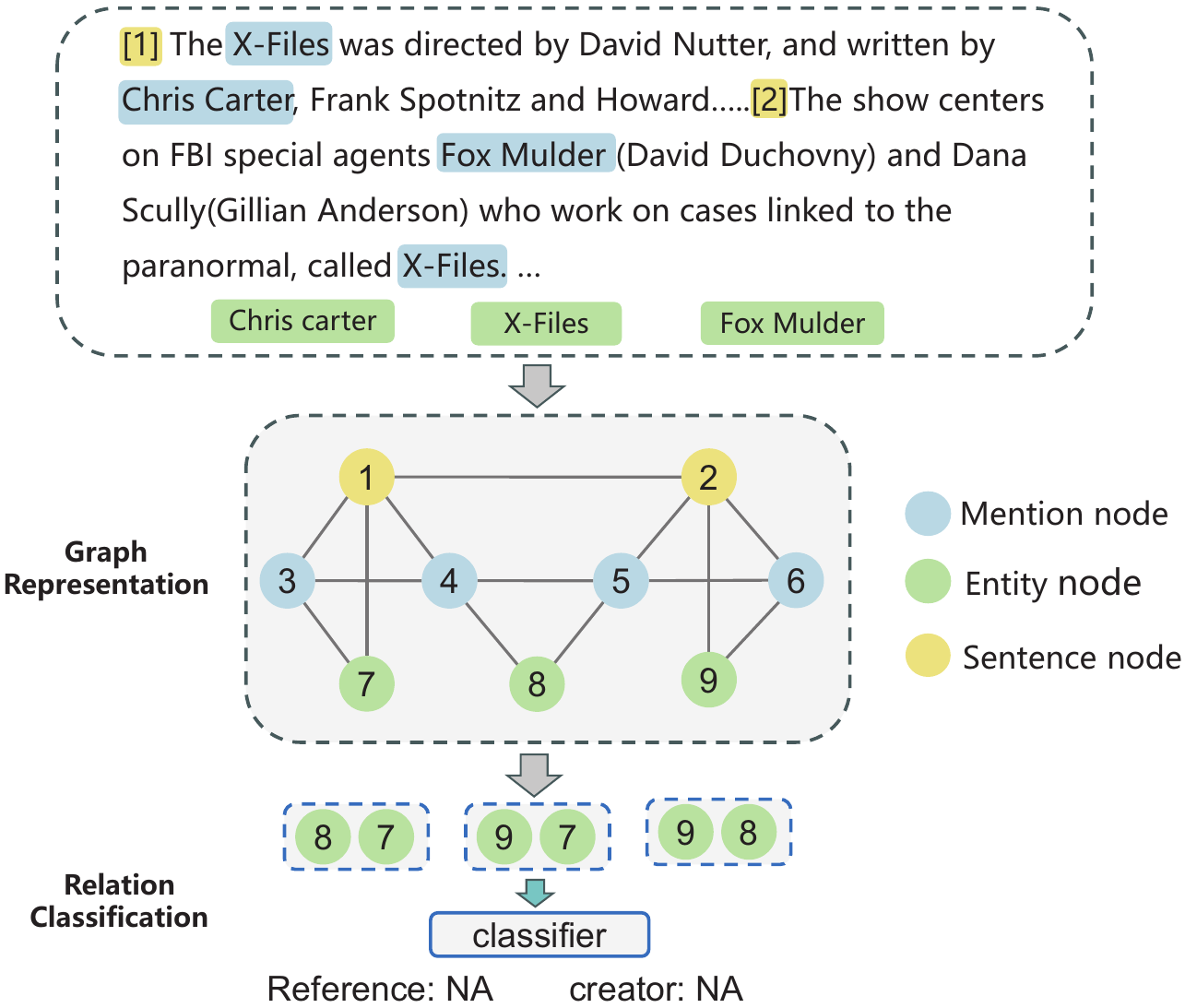}
            \caption{Relation Extraction.}
            \label{fig:RE_a}
        \end{subfigure}
        
        \vspace{5pt} 
        \begin{subfigure}[b]{\linewidth}            \includegraphics[width=0.85\linewidth]{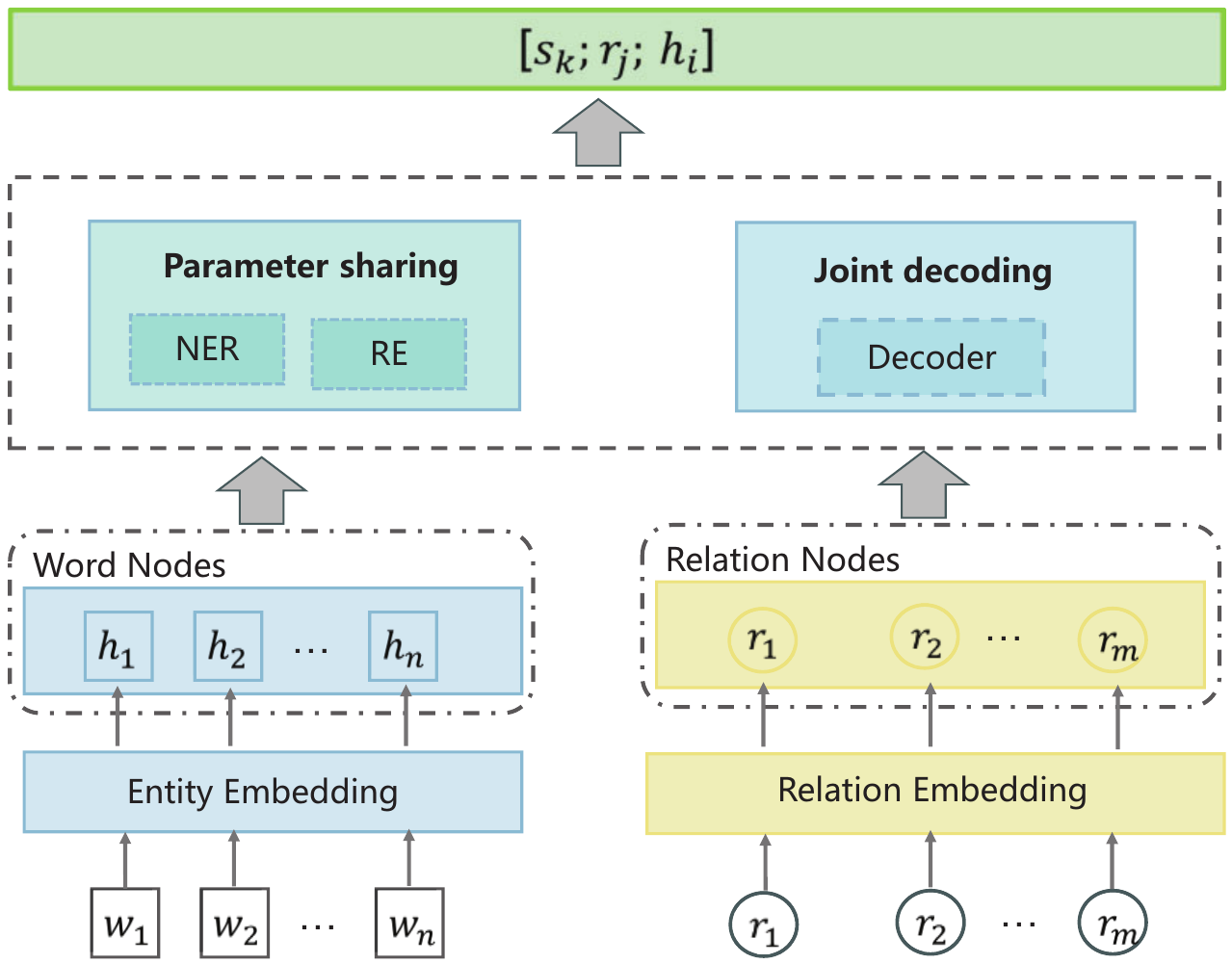}
            \caption{Joint Entity and Relation Extraction.}
            \label{fig:RE_b}
        \end{subfigure}
    \end{minipage}
    \Description{Different knowledge extraction tasks: Named Entity Recognition, Relation Extraction, and Joint Extraction.}
    \caption{Different knowledge extraction tasks: Named Entity Recognition, Relation Extraction, and Joint Extraction.}
    \label{fig:FIG4}
\end{figure}

\subsubsection{Named Entity Recognition}
\ 
\newline NER aims to identify and classify specific entities in text, such as names of persons, locations, and organizations. Graph neural network-based NER methods effectively model deep semantic dependencies between words by transforming text sequences into nonlinear graph topologies  \cite{sui2019leverage,ding2019neural,wang2022polymorphic}. In text-structure-based modeling, M-DGNN \cite{gu2024enhanced} and DGNN \cite{liu2022learning} focus on leveraging prior syntactic knowledge by constructing dependency syntax graphs or co-occurrence graphs, utilizing GCN's aggregation capabilities to capture long-range semantic associations. In contrast, GNNer \cite{zaratiana2022gnner} constructs text span-symbol graphs for non-nested entities, employing a combination of GCN and GAT to reduce information redundancy caused by overlapping boundaries. Compared with the above methods, RWGNN \cite{chen2023randomly} introduces a random edge mechanism to automatically generate multi-directional connection patterns, which solves the problem of existing GNN models relying on fixed graph structures.

Due to the blurred boundaries and ambiguity of entities, many studies integrate external knowledge. The core challenge in composition lies in how to handle heterogeneous nodes. LGN \cite{gui2019lexicon}, MCGAT \cite{zhao2021multi}, and PGAT \cite{wang2022polymorphic} commonly adopt character-word polymorphism graph topology, connecting fragmented characters through identified lexical units as intermediary nodes. Since external knowledge often introduces noise, these models consistently incorporate attention mechanisms, dynamically adjusting the contribution weights between nodes through attention. KaNa \cite{nie2021knowledge} and BAC-GNN-CRF \cite{xu2024semantic} further elevated the compositional dimension to the entity type level and paragraph heading level, achieving multi-level integration of local vocabulary and macro-context with the assistance of GAT.

Heterogeneous Graph Neural Networks \cite{SUN2025130369} can handle graph structures with multiple types of nodes and edges, making them suitable for complex NER tasks such as nested entity recognition and cross-domain NER. 
For nested entity recognition or cross-modal tasks, the architecture design places greater emphasis on logical reasoning and alignment. Trigger-GNN \cite{sui2022trigger} and HGFNER \cite{zhang2024chinese} simulate structured reasoning processes by constructing entity trigger word graphs or space-syntax heterogeneous graphs. In the multimodal named entity recognition (MNER) domain, RGCN \cite{zhao2022learning} and MOUSING \cite{lu2024few} construct intricate cross-modal interaction graphs. To address feature biases across modalities, the models employ R-GCN or causal intervention mechanisms.

\subsubsection{Relation Extraction}
\ 
\newline The goal of relation extraction \cite{DBLP:conf/emnlp/HaoYH21} is to automatically identify semantic relationships between entities from text. For example, in the sentence "Einstein was born in Germany," the relation extraction needs to identify the semantic relation "place of birth" between the entities "Einstein" (entity 1) and "Germany" (entity 2). By modeling sentence topology as graphs, GNNs can define semantic distances between words more precisely than sequence models and better capture complex relationships among entities. 
 
In sentence-level tasks, structural augmentation is often performed based on syntactic topology. For example, C-GCN \cite{DBLP:conf/emnlp/Zhang0M18} and LSTAGCN \cite{sun2020relation} leverage GCNs to perform preliminary aggregation of structural information by constructing weighted dependency syntax graphs. To further filter out irrelevant nodes in the syntax tree, AGGCN \cite{DBLP:conf/acl/GuoZL19}, DAGCN \cite{li2021improve}, and A-GCN \cite{DBLP:conf/acl/TianCSW20}incorporate attention mechanisms that transform dependency trees into fully connected weighted graphs, enabling the model to dynamically focus on key semantic paths.

In document-level tasks, capturing long-range dependencies across sentences is crucial. EoG \cite{DBLP:conf/emnlp/ChristopoulouMA19}, HeterGSAN \cite{xu2021document}, and GLRE \cite{DBLP:conf/emnlp/WangHCS20} commonly adopt hierarchical heterogeneous graphs based on the “Mention-Entity-Sentence” architecture, utilizing GAT or path-enhanced GNNs to enable cross-sentence feature propagation. GAIN \cite{DBLP:conf/emnlp/ZengXCL20}and DUALGRAPH \cite{DBLP:journals/tnn/LiFC24} support complex multi-hop reasoning by constructing a dual-layer graph structure (mention graph and entity graph) and leveraging GCNs to align information across different granularities. To enhance reasoning depth, RECON \cite{DBLP:conf/www/BastosN0MSHK21} enables models to focus more on genuinely existing entity relationships through knowledge graph context and path reconstruction. 

To break free from reliance on predefined syntax trees, some research has shifted toward automatically learning compositions. GDPNet \cite{xue2021gdpnet} utilizes the Gaussian Graph Generator (GGG) to generate edges for multi-view graph structures and employs graph pooling to prune redundant information. GP-GNN \cite{DBLP:conf/acl/ZhuLLFCS19} further enhances compositional flexibility by dynamically generating the adjacency matrix parameters used in its message passing process from textual sentences. LSR \cite{DBLP:conf/acl/NanGSL20} automatically discovers association paths between entities during training by constructing a Latent Dynamic Graph.

\subsubsection{Joint Extraction}
\
\newline Joint extraction \cite{zhao2021unified} aims to simultaneously identify entities and their relationships within text, thereby overcoming the cumulative error issues inherent in traditional pipeline approaches. GNNs represent entities and relations in text by constructing graph structures, effectively capturing their complex interactions and rich contextual information\cite{sun2019joint, carbonell2021named}. 

Early research primarily focused on converting the syntactic structure of text into graph topology. Both GraphRel \cite{fu2019graphrel} and CPJE \cite{wang2022conditional} employ dependency syntax graphs combined with GCNs for modeling. Among these, GraphRel enhances interactions between entities through relation-weighted GCNs, while CPJE goes a step further by mitigating cascading errors in biomedical event extraction through consideration of trigger word probability distributions. Additionally, TAG \cite{zhang2023novel} and DGAT \cite{zhao2024joint} capture word-level semantics by constructing mention graphs and word graphs, respectively.

To capture deeper semantic meanings, researchers have begun constructing heterogeneous graphs incorporating multiple node types. ESEI \cite{li2023joint} and RIFRE \cite{zhao2021representation} have all moved away from single-entity graphs toward constructing heterogeneous graphs composed of entities, relations, and even entity types and text spans. For instance, ESEI leverages GAT to learn higher-order interactions among entities by constructing span graphs and type graphs; RIFRE and its biological variant Bio-RIFRE \cite{esmail2022chemical} focus on modeling entity-relationship graphs, enhancing triplet extraction performance through explicit modeling of relationship nodes.

When confronted with nested relationships and more complex semantic contexts, single-view or low-level graph structures reveal their limitations. HGERE \cite{DBLP:conf/emnlp/0001YLT23} introduces HyperGraph Neural Networks (HyperGNN), which effectively address error propagation issues in nested relationships by constructing hypergraphs to capture higher-order interactions among multiple entities and their relationships. In multi-view fusion, both MVGNAS \cite{al2022multi} and BanglaAutoKG \cite{DBLP:conf/coling/WasiRIC24} adopt a hybrid architecture combining GCN and GAT. MVGNAS enhances the representation of the Span graph through multi-view learning, while BanglaAutoKG combines multilingual large language models with graph-based semantic filtering to automate the knowledge graph construction process. These models demonstrate that integrating the structural modeling capabilities of GNNs with the semantic representation capabilities of pre-trained language models represents the current mainstream approach for tackling complex extraction tasks.

\subsection{Knowledge Fusion}
Knowledge fusion aims to eliminate heterogeneity among different knowledge sources, primarily encompassing two core tasks: entity alignment and entity linking. As summarized in Table \ref{tab:Knowledge Fusion}, existing research on entity alignment primarily focuses on modeling challenges related to global semantics, complex topology, and domain-specific aspects. In entity linking tasks, research emphasis lies on relational modeling, handling structural heterogeneity, and achieving robustness in complex scenarios. And the synthesis of representative methodologies presented in Table \ref{tab:Knowledge Fusion} is conducted through the lens of graph construction and model architecture.
\subsubsection{Entity Alignment}
\ 
\newline Entity alignment aims to identify entities from different knowledge graphs that refer to the same real-world object \cite{DBLP:conf/ijcai/SunHZQ18}. 
Traditional entity alignment methods primarily rely on symbolic logic rules \cite{DBLP:journals/pvldb/SuchanekAS11} or similarity-based \cite{raimond2008automatic} heuristic algorithms. However, these approaches often struggle with computational inefficiency and limited generalization capabilities when dealing with large-scale and complex-structured knowledge graphs. In recent years, embedding-based methods have gradually become mainstream, particularly with the introduction of GNNs, which provide powerful modeling tools for entity alignment. GNNs can aggregate neighborhood information of entities through message-passing mechanisms, generating low-dimensional vector representations, thereby measuring entity similarity in a unified embedding space \cite{xu2020coordinated}. Early GNN alignment models often focused on topological structures while neglecting the semantic meaning of relations. To address this issue, RNM \cite{zhu2021relation}, MRAEA \cite{mao2020mraea}, HGCN \cite{DBLP:conf/emnlp/WuLFWZ19}, and VR-GCN \cite{ye2019vectorized}  collectively explored methods to enhance representations through orthogonal transformations, relation-aware attention, or entity-relation joint learning. RDGCN \cite{DBLP:conf/ijcai/WuLF0Y019} further strengthens interactions between relations by constructing a Dual Relation Graph, while EPEA \cite{wang2020knowledge} utilizes an edge-aware attention mechanism to propagate features within a Pair-wise Connectivity Graph.

Structural heterogeneity between knowledge graphs poses a significant challenge for entity alignment. AliNet \cite{sun2020knowledge} addresses this issue by introducing gated multi-hop neighborhood aggregation and employing attention mechanisms to filter distant neighbors. NMN \cite{DBLP:conf/acl/WuLFWZ20}, on the other hand, jointly computes neighborhood differences between sampled subgraph pairs through cross-graph attention mechanisms. MuGNN \cite{DBLP:conf/acl/CaoLLLLC19} employs a multi-channel graph neural network to coordinate structural asymmetry, while SSP \cite{wong2020global} combines global structure with local semantics for attribute enhancement. To compensate for the lack of structural information, researchers introduced diversified features. At the attribute dimension, AttrGNN \cite{DBLP:conf/emnlp/LiuCPLC20}  and ACK-MMEA \cite{li2023attribute}  respectively construct attribute graphs and text/image graphs, utilizing GAT to address context bias. EVA \cite{liu2021visual} uses visual semantics as the alignment pivot. Additionally, HMAN \cite{DBLP:conf/emnlp/YangZSLLS19} integrates the semantic information from cross-lingual pre-trained models, significantly improving alignment accuracy in cross-lingual scenarios. Temporal-aware methods, such as TEA \cite{liu2023tea}, STEA \cite{DBLP:conf/coling/CaiMMYZL22}, TEA-GNN \cite{DBLP:journals/corr/abs-2203-02150}, and DualMatch \cite{liu2023unsupervised}, capture entity evolution patterns through temporal graph encoding and dynamic context clustering.

Research on complex scenarios has also made notable progress.In the context of dynamic KG alignment, the DINGAL \cite{yan2021dynamic} series propose topology-invariant gating and incremental update strategies to achieve efficient dynamic embedding updates. LargeGNN \cite{xin2022large} employs a subgraph partitioning strategy based on centrality, effectively reducing computational complexity for ultra-large-scale graphs. To address missing labels, ActiveEA \cite{DBLP:conf/emnlp/LiuSZHZ21} introduces active learning to optimize sampling; Lambda \cite{yinlambda} employs contrastive learning specifically to handle unlabeled dangling entities. 

\begin{figure}[!htbp]
    \centering
    \begin{subfigure}[b]{0.49\textwidth}        \includegraphics[width=\linewidth, height=4.7cm]{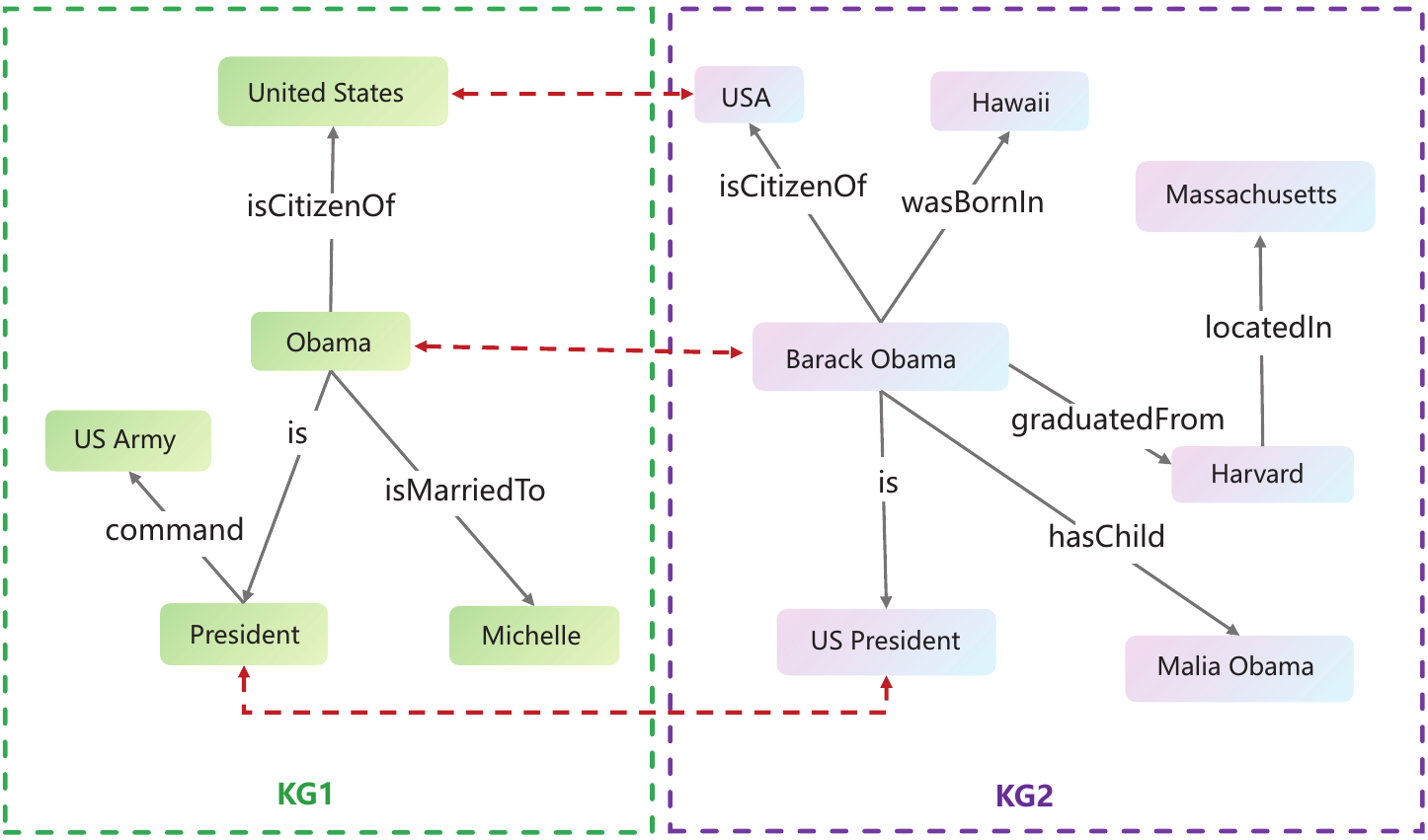}
        \caption{Entity alignment}
        \label{fig:Entity_Alignment}
    \end{subfigure}
    \begin{subfigure}[b]{0.49\textwidth}        \includegraphics[width=\linewidth, height=4.6cm]{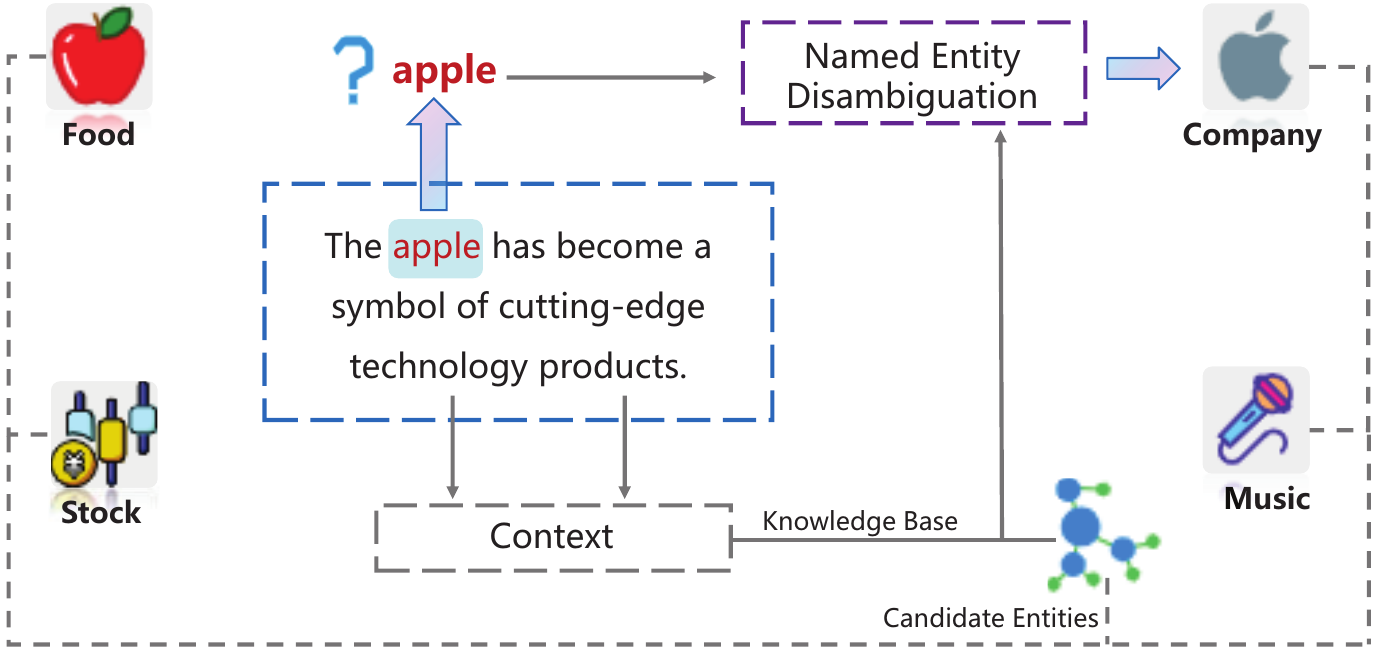}
        \caption{Entity linking}
        \label{fig:Entity_Linking}
    \end{subfigure}
    \Description{Entity alignment and linking}
    \caption{Entity alignment and linking: (a) Entity alignment, identifying and matching equivalent entities in two knowledge graphs. (b) Entity linking, resolving ambiguous entities through context within a single graph (e.g., "Apple" as a company or music).} \label{fig:Entity_Alignment_linking}
\end{figure}

\subsubsection{Entity Linking}
\ 
\newline Entity linking, also known as entity disambiguation. Its goal is to associate entity mentions in text with unique entities in a knowledge base. 
General GNN-based entity linking frameworks primarily leverage graph modeling to capture the global semantic consistency between entity mentions and knowledge base entities\cite{DBLP:conf/bigdataconf/KrivosheevMAS23,DBLP:journals/tnn/ZhouYWGB24}. 

Early GNN entity linking research primarily focused on overcoming the limitations of local context by constructing graph structures to capture document-level global semantics. GNED \cite{DBLP:journals/kbs/HuDSSL20} pioneered the construction of an “entity-word” heterogeneous graph, leveraging GCNs to generate enhanced entity embeddings that overcome traditional models' excessive reliance on local features. Subsequent research has further enhanced the flexibility of graph structures. DGCN \cite{DBLP:conf/www/WuZMGSH20} introduces a dynamic GCN architecture capable of adaptively adjusting the graph structure based on document context. SGEL \cite{DBLP:conf/www/FangCLZLW20} combines sequence models with GAT, enabling deep semantic dependency mining by sequentially encoding contextual references. CoGCN \cite{DBLP:journals/es/JiaCSD21} focuses on interaction mechanisms by combining co-attention with GCN to establish associations between mention contexts and candidate entities, significantly enhancing the automation of feature extraction.

Entity link aggregation emphasizes consistency among multiple mentions within a document, with research focusing on how to perform joint inference through complex graph topologies. EL-Graph \cite{DBLP:conf/nlpcc/WangXK23} employs a joint subgraph-based approach that utilizes GAT to filter key mentions and construct local subgraphs, thereby performing joint disambiguation within these subgraphs to effectively enhance link accuracy. HEGEL \cite{DBLP:journals/dint/ChenWFZ22} elevates heterogeneity to a higher dimension by integrating diverse knowledge sources such as semantics and neighborhood information. It constructs a heterogeneous “mention-entity-keyword” graph, demonstrating the critical role of incorporating multi-source prior knowledge in enhancing global consistency modeling.

Researchers typically design customized GNN models for specific domains. GNN-RE \cite{yin2023two} addresses Building Information Modeling (BIM) retrieval by integrating IFC ontology knowledge with dependency syntax graphs. It employs GAT to map natural language queries into structured retrieval statements, thereby resolving the challenge of domain-specific term mapping. Regarding the complex semantic associations between mentions and entities in medical texts, ED-GNN \cite{DBLP:conf/sigmod/Vretinaris0EQO21} and MMR-FEK \cite{DBLP:conf/coling/LuZPGW24} offer distinct solutions. The former combines the strengths of GCN and GAT, focusing on optimizing the negative sampling strategy to enhance discriminative power. The latter reinforces the interaction between mention representations and fine-grained entity knowledge through a bidirectional attention mechanism. The commonality between the two lies in their use of multiple views.

\section{Knowledge Graph Embedding}
\label{KGE}
Knowledge Graph Embedding (KGE) aims to map entities and relations in knowledge graphs into low-dimensional continuous vector spaces. Through such embedding representations, the complex relationships in knowledge graphs are simplified and quantified. These embeddings provide an efficient computational foundation for downstream tasks, such as knowledge reasoning, entity classification, relation extraction, and link prediction. Early KGE methods primarily relied on the translation assumption, which models the interactions between entities and relations through linear transformations in vector spaces\cite{DBLP:conf/nips/BordesUGWY13,DBLP:conf/aaai/WangZFC14,DBLP:conf/aaai/LinLSLZ15}. On the other hand, models like DistMult \cite{DBLP:journals/corr/YangYHGD14a} and ComplEx \cite{DBLP:conf/icml/TrouillonWRGB16} adopted multiplicative models or tensor decomposition models, utilizing multiplicative scoring functions to evaluate the plausibility of triples, thereby capturing intricate interactions between entities and relations. Despite the notable success of these embedding methods in link prediction, they share a common limitation: the neglect of local neighborhood structures in knowledge graphs. In contrast, GNNs can effectively learn node representations by recursively aggregating and transforming embeddings from neighboring nodes, thereby better capturing the local structural information of knowledge graphs. 
In this section, we introduce GNN-based KGE methods for different types of knowledge graphs, including static knowledge graphs, temporal knowledge graphs, and complex knowledge graphs. 
For static graphs, existing methods primarily enhance the richness and quality of representation learning by exploring novel GNN architectures or performing parameter optimization. In contrast, research on dynamic graphs has shifted its focus to capturing the temporal evolution of sequential information, encompassing parallel research directions such as inductive learning, enhanced temporal modeling, and time-aware attention mechanisms. For complex graphs incorporating multi-modal information or diverse relationships, researchers concentrate on hyper-relational modeling and multi-modal feature integration. Table \ref{tab:KGE} provides a refined distillation of these representative methods through a dual analysis of their underlying graph construction strategies and specific GNN architectures.

\begin{figure}[!htbp]
    \centering
    \begin{minipage}[t]{0.49\linewidth}  
        \begin{subfigure}[t]{\linewidth}
            \includegraphics[width=\linewidth]{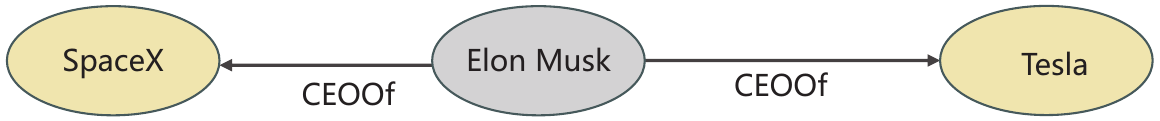}
            \caption{Triple-shaped KG}
            \label{fig:triple}
        \end{subfigure}
        
        \begin{subfigure}[t]{\linewidth}
            \includegraphics[width=\linewidth]{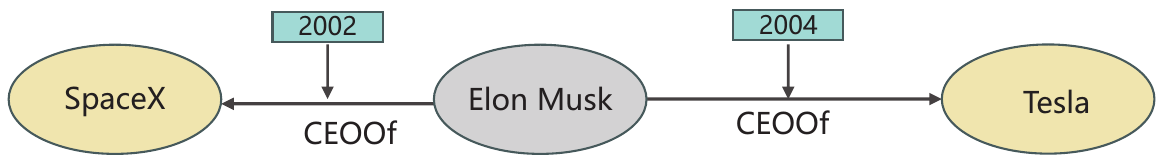}
            \caption{Temporal KG}
            \label{fig:temporal}
        \end{subfigure}
        
        \begin{subfigure}[t]{\linewidth}
            \includegraphics[width=\linewidth]{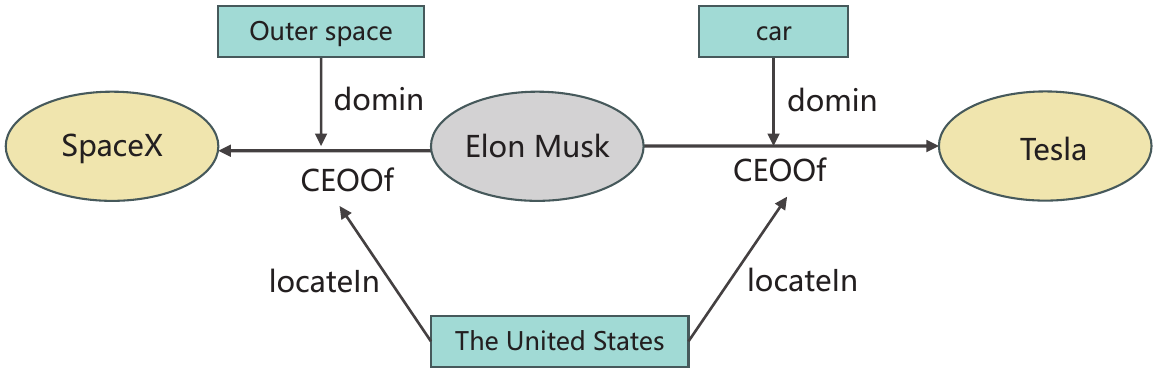}
            \caption{Hyper-relational KG}
            \label{fig:hyper}
        \end{subfigure}
    \end{minipage}
    \hfill
    \begin{minipage}[t]{0.49\linewidth}
        \vspace*{-0.5cm} 
        \centering
        \begin{subfigure}[t]{\linewidth}
            \includegraphics[width=\linewidth ,height=12cm, keepaspectratio]{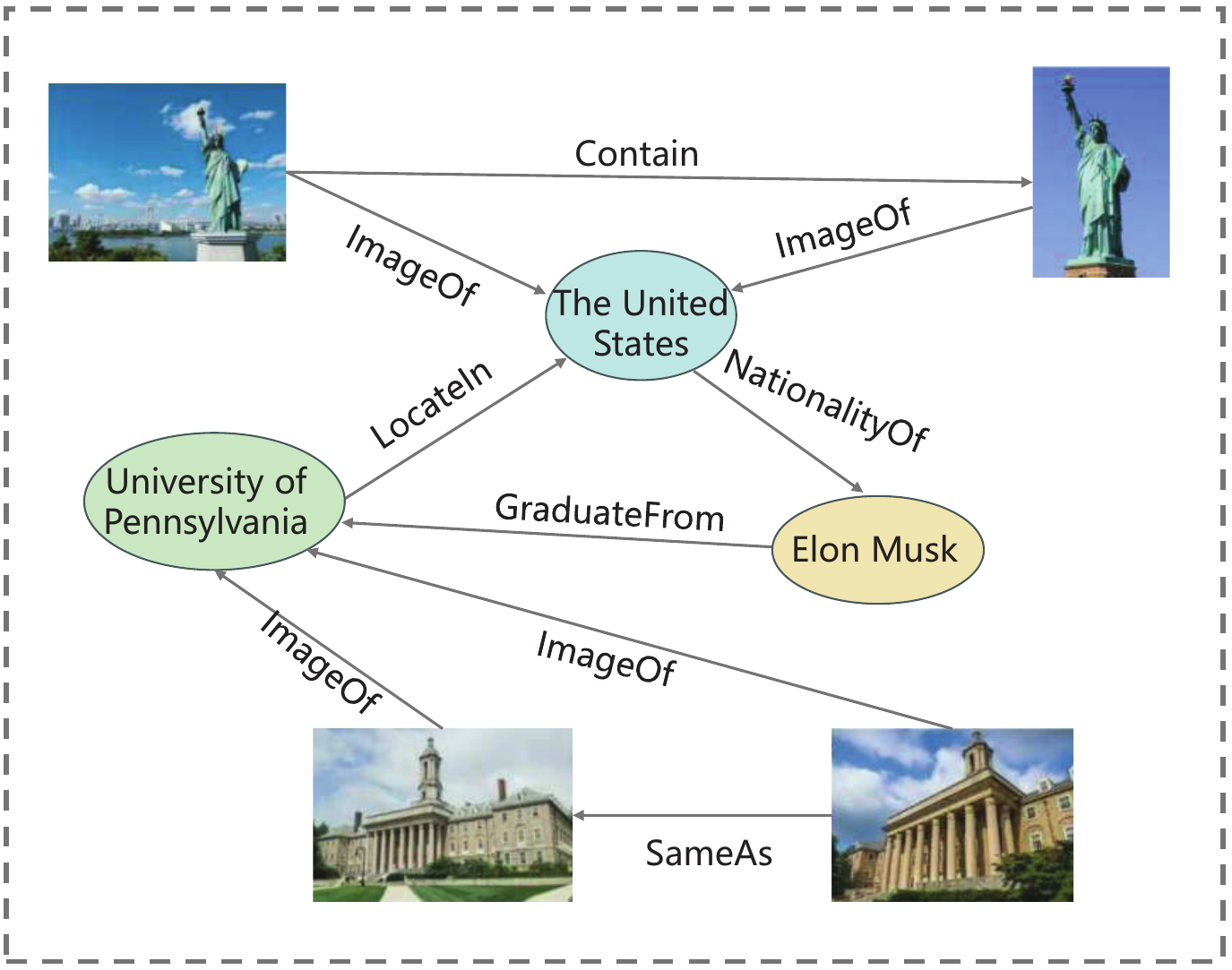}
            \caption{Multimodal KG} 
            \label{fig:nary}
        \end{subfigure}
        \vspace*{\fill}
    \end{minipage}
    \Description{A list of shapes of KGs}
\caption{A list of shapes of KGs (including Static KG, temporal KG, hyper-relation KG, and multimodal KG).}
\label{fig:kg_structures}
\end{figure}

\label{Static Knowledge Graph}
\subsection{Static Knowledge Graph}

Static KGC research focuses on leveraging GNNs to capture complex topological structures and enhance parameter efficiency. Early methods such as R-GCN \cite{schlichtkrull2018modeling} laid the foundation for processing multi-relational data by introducing relationship-specific transformation matrices. Subsequent studies such as M-GNN \cite{wang2019robust} and MSeaKG \cite{di2023message} further expanded the perspective from single graphs to hierarchical graphs and more flexible message-passing search spaces. SE-GNN \cite{li2022does} introduces a semantically aware architecture that dynamically assigns neighborhood weights through the GAT mechanism, significantly enhancing the model's inference robustness.

As knowledge graphs grow in scale, some researchers have begun shifting their focus to model efficiency and parameter optimization. Both EARL \cite{chen2023entity} and decentRL \cite{guo2024distributed} adopt an entity-agnostic approach. The former reduces storage through subset embedding, while the latter utilizes decentralized attention to distribute semantic information. HKGN \cite{liu2022heterogeneous}, on the other hand, introduces hypernetworks to generate relation-specific parameters, improving parameter efficiency while ensuring strong model performance.

\subsection{Dynamic Knowledge Graph}
\label{Dynamic Knowledge Graph}
As time progresses, knowledge continuously develops, updates, and evolves. The dynamic and time-sensitive nature of knowledge has become increasingly important in real-world applications. Dynamic graphs not only capture changes in knowledge but also reflect its evolution over time, allowing for continuous updates and expansion. The core challenge of Dynamic KG lies in addressing the immediate updates of knowledge (inductive nature) and the evolutionary patterns of the temporal dimension (temporal nature).

Inductive reasoning for structural changes aims to break free from the reliance of transductive methods on predefined entities, enabling models to generalize to unseen entities. Such methods typically employ GAT variants as their underlying architecture, utilizing attention mechanisms to adaptively aggregate information from heterogeneous neighborhoods. GraIL \cite{teru2020inductive}, SNRI \cite{DBLP:conf/ijcai/XuZHCY22}, and CoMPILE \cite{DBLP:conf/aaai/MaiZY021} no longer rely on predefined entity embeddings but instead constitute a class of typical paradigms based on local subgraphs. Although these methods all utilize closed subgraphs, their focuses differ. GraIL concentrates on local topology, SNRI incorporates global neighborhood information through mutual information maximization, while CoMPILE optimizes message passing between nodes and edges. Furthermore, InGram \cite{DBLP:conf/icml/LeeCW23} achieves generalized processing of new relationships, while DEKG-ILP \cite{DBLP:conf/icde/ZhangWYZCZ23} introduces contrastive learning to capture global semantic features.

Temporal knowledge graphs are a special type of dynamic knowledge graph, with the core focus on explicitly incorporating the time dimension into knowledge representation and reasoning processes. RE-GCN \cite{DBLP:conf/sigir/LiJLGGSWC21} leverages relation-aware GCNs to capture structural dependencies within snapshots, combined with gated recurrent components for historical evolution modeling. GTRL \cite{DBLP:journals/tkde/TangC24} and TeMP \cite{DBLP:conf/emnlp/WuCCH20} extend this approach. DACHA \cite{DBLP:journals/tkdd/ChenTCQLZ22} employs a dual-graph convolution architecture to model the original graph and edge graph separately, thereby explicitly capturing both local and global temporal dependencies. These approaches share the commonality of leveraging the robust structural representation capabilities of GCNs and their variants to transmit temporal information through stacked time steps. Another category of research focuses on leveraging time-sensitive attention mechanisms to capture complex temporal relationships. xERTE \cite{DBLP:conf/iclr/HanCMT21}, TARGAT \cite{DBLP:journals/taslp/XieZ0ZH23} and LogCL \cite{DBLP:conf/icde/ChenWWZCLL24} introduced time-aware attention. These methods no longer treat time as a simple sequence but dynamically generate time-sensitive transformation matrices. LogCL and xERTE further integrate contrastive learning and reverse representation update mechanisms respectively, enhancing the model's ability to extract key historical information relevant to queries.

\subsection{Complex knowledge graph}
\label{Complex knowledge graph} 
With the continuous advancement of knowledge graph technologies and its widespread adoption in practical applications, the representation of knowledge has evolved from traditional simple triplets to more complex structures. Notable examples include multimodal knowledge graphs and hyper-relational knowledge graphs.

Knowledge hypergraphs, as an extension of traditional knowledge graphs, connect multiple nodes through hyperedges, offering a more accurate representation of multi-dimensional relationships. Early hypergraph embedding methods \cite{DBLP:conf/ijcai/WenLMCZ16, DBLP:conf/ijcai/FatemiTV020, DBLP:conf/acl/GuanJGWC20} often treated hyperedges as independent entities, neglecting the structural information shared between hyperedges that share common nodes. To address the aforementioned issues, HeteHG-VAE \cite{DBLP:journals/pami/FanZWLZGD22} and H2GNN \cite{DBLP:journals/corr/abs-2412-12158} introduced the Heterogeneous Hypergraph Neural Network (HGNN) architecture. Compared to the representational limitations that may arise from HeteHG-VAE's reliance on class-based methods, H2GNN directly embeds hyper-edge information through a star-shaped message passing mechanism, reducing modeling complexity while preserving hierarchical structure. DHGE \cite{DBLP:conf/aaai/LuoETZYW23} takes this further by combining HGNN with graph transformers, proposing a dual-view framework comprising instance views and ontology views. Regarding additional information such as qualifiers present in hyper-relational knowledge graphs, STARE \cite{DBLP:conf/emnlp/GalkinTMUL20} focuses on synchronously processing the main triplet and any number of qualifiers using a specific message-passing framework. In contrast, QUAD \cite{DBLP:conf/asunam/ShomerJL0L23} introduces bidirectional aggregation operators to enhance the depth of interaction between the main triplet and qualifiers.

Multimodal KGs overcome the limitations of traditional symbolic representations by integrating text, image, audio, and other data types. Current state-of-the-art approaches commonly employ GAT to dynamically assign weights to different modalities. CMGNN \cite{DBLP:journals/tkde/FangZHWX23} employs a contrastive learning mechanism to enhance entity representations by extracting shared features from multimodal content and higher-order connection structures. In contrast, HRGAT \cite{DBLP:journals/tomccap/LiangZZ023} combines low-rank multimodal fusion with GAT, using attention mechanisms to finely tune interactions between modalities.

\section{Knowledge Reasoning}
Knowledge reasoning aims to uncover implicit logical connections or predict future trends by leveraging existing knowledge. Based on differences in inference objectives and application contexts, we categorize this field into Inference for Completion and Inference for Prediction. For completion reasoning, existing research primarily focuses on enhancing the logical inference capabilities of static knowledge through novel graph construction, enhanced attention mechanisms, multi-source fusion, and causal inference. In predictive reasoning tasks, the research focus has shifted toward modeling temporal dynamics. This branch is primarily categorized into two major methods based on underlying architectures: GCN-based and GAT-based approaches. We systematically synthesize the representative approaches summarized in Table \ref{tab:Knowledge Reasoning} by examining the two pivotal perspectives of graph construction strategies and GNN architectures.
\label{KR}
 \subsection{Inference for Completion}
\begin{figure}[!htbp]
    \centering
    \begin{subfigure}[b]{0.4\textwidth}
        \includegraphics[width=\linewidth]{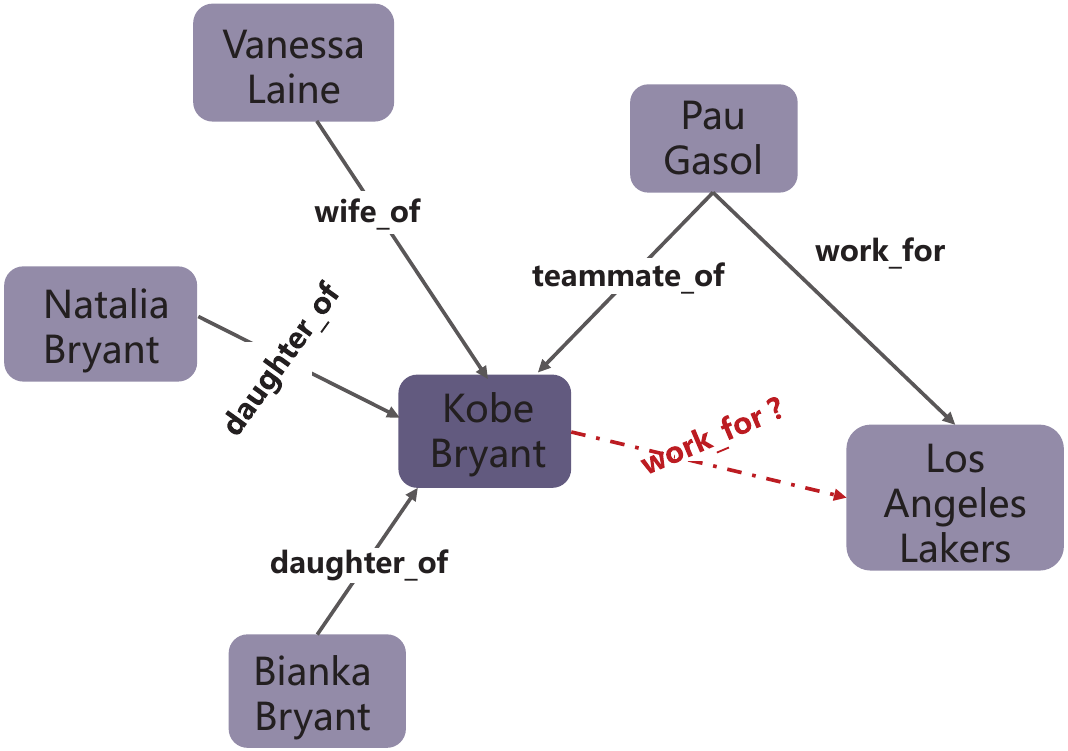}
        \caption{Inference for Completion}
        \label{fig:KGC}
    \end{subfigure}
    \begin{subfigure}[b]{0.56\textwidth}
        \includegraphics[width=\linewidth]{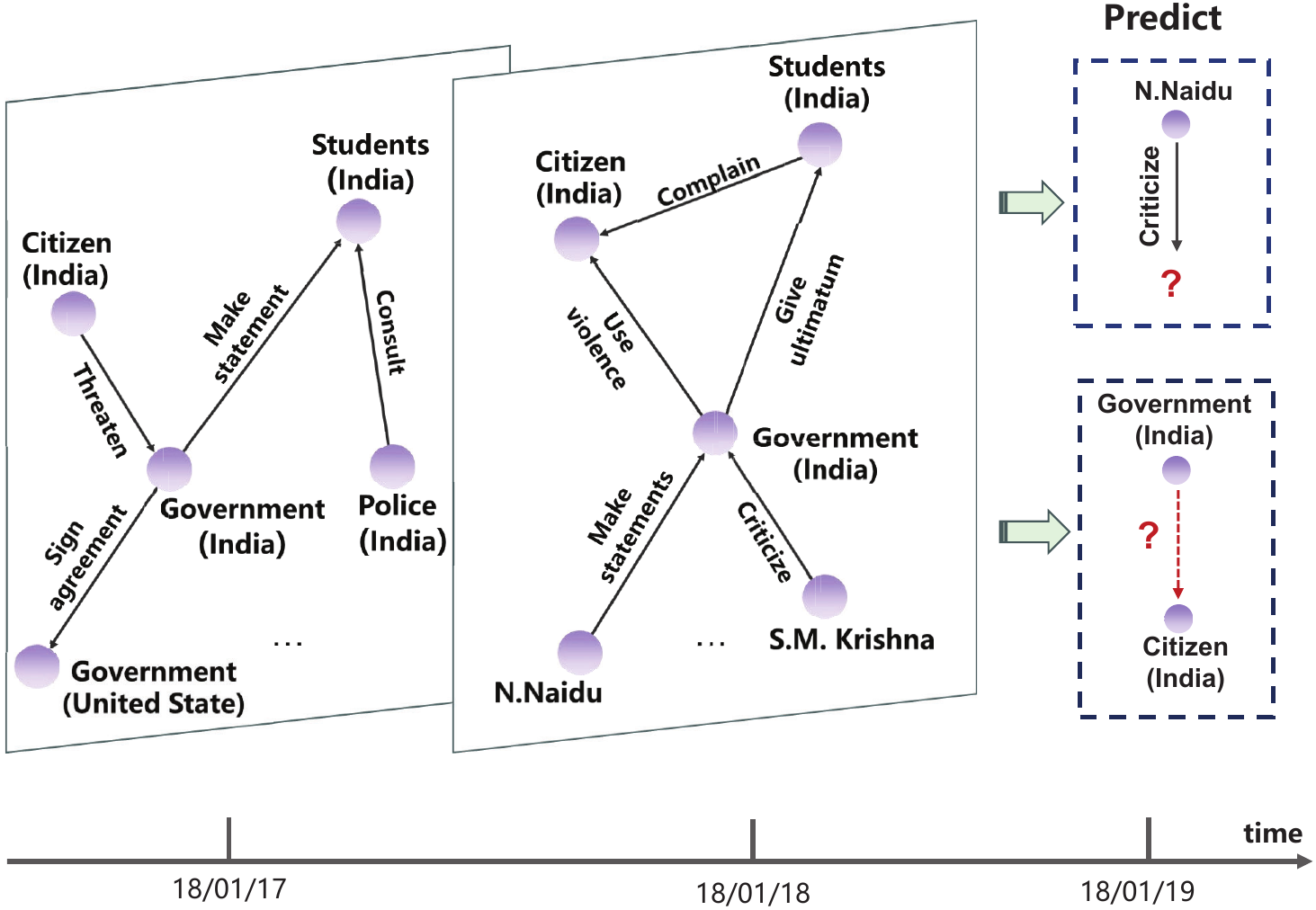}
        \caption{Inference for Prediction}
        \label{fig:TKGC}
    \end{subfigure}
    \Description{Inference for completion and prediction}
    \caption{Inference for completion and prediction: (a) Example of knowledge graph completion. The dashed arrows are the relationships between two people (nodes) that need to be inferred (predicted triples); (b) Explanation of temporal reasoning on TKG. We present two subgraphs from KG on timestamps 18/01/17 and 18/01/18, respectively.}
    \label{fig:KGC2TKGC}
\end{figure}

KGs are widely used in various artificial intelligence tasks as a graphical data model representing structured knowledge through nodes (entities) and edges (relationships). However, one issue is that KGs are often incomplete, meaning they lack many valid triples \cite{DBLP:journals/kbs/ShenZC22}. This incompleteness poses a significant challenge for applications relying on knowledge graphs. To address this issue, knowledge graph completion has become a key research focus. The core task of KG completion is to predict and fill in the missing triples by reasoning over the known triples in the existing knowledge graph, shown in Figure \ref{fig:KGC}. Many KG embedding models have been proposed to predict whether triples not present in the KG are likely valid, such as TransE \cite{DBLP:conf/nips/BordesUGWY13}, DistMult \cite{DBLP:journals/corr/YangYHGD14a}, ComplEx \cite{DBLP:conf/icml/TrouillonWRGB16}, and QuatE \cite{DBLP:conf/nips/0007TYL19}. These models aim to learn vector representations of entities and relationships, using different scoring functions to assess the validity of triples, ensuring that valid triples receive higher scores than invalid ones.

In recent years, the encoder-decoder architecture based on GNNs has become the mainstream approach in KGC \cite{chen2022knowledge, DBLP:conf/coling/AnilGIS24}. These methods capture complex topological structures through GNN encoders and perform inference using scoring functions. Current improvements primarily focus on innovative graph structures, mechanism enhancement, semantic fusion, and causal reasoning.

To overcome the limitations of traditional topological structures, researchers no longer confine themselves to primitive entity graphs. WGE \cite{DBLP:conf/esws/TongNPN23} and NoGE \cite{DBLP:conf/wsdm/NguyenTPN22} attempt to capture higher-order associations by constructing entity-relation graphs or weighted entity graphs, transforming relations into nodes or weights, thereby leveraging GCNs. To address the heterogeneity and sparsity of knowledge graphs, INDIGO \cite{DBLP:conf/nips/LiuGHK21} introduces Pairwise Encoding to enable inductive reasoning, while M2GNN \cite{DBLP:conf/www/WangWSWNAXYC21} and LR-GCN \cite{DBLP:journals/fcsc/HeLCWZQ25} enhance the model's representation capabilities for complex geometric structures and sparse graphs through multi-curvature space mapping and long-range dependency capture, respectively. Additionally, TAGNet \cite{DBLP:conf/emnlp/Shomer0LWAT23} significantly improves path aggregation efficiency via a truncated propagation mechanism.

To address the challenge of distinguishing importance when aggregating neighborhood information in GNNs, introducing more sophisticated attention mechanisms has become crucial for enhancing performance. A common feature of such studies (e.g., RGHAT \cite{DBLP:conf/aaai/ZhangZZ0XH20}, GATH \cite{DBLP:journals/tkdd/WeiSY24}, and MRGAT \cite{DBLP:journals/nn/DaiWZLC22}) is the extension of single-level attention to dual or multi-level mechanisms at entity and relationship levels, enabling more effective balancing of heterogeneous connections. DisenKGAT \cite{DBLP:conf/cikm/WuSCCLZW021} further achieves decoupling between micro- and macro-structures, while DRR-GAT \cite{DBLP:journals/eswa/ZhangZGPNW23} focuses on addressing global attribute loss through dynamic representations. Notably, KRACL \cite{DBLP:conf/www/TanCFZZLL23} combines attention networks with contrastive learning, leveraging self-supervised signals to mitigate severe data sparsity challenges.

Beyond topological structures, incorporating multi-source information can significantly enrich entity representations. DMoG \cite{DBLP:conf/coling/SongHZG00022} and D-GNN \cite{wu2023graph} address the lack of structural information by integrating ontology graphs and text graphs; LSMGA \cite{DBLP:conf/acl/TangZZZ23} tackles alignment dependency issues in multilingual scenarios through language-sensitive mechanisms. Another trend involves enhancing logical rules. For instance, AR-KGAT \cite{DBLP:journals/kbs/ZhangHT22} incorporates logical rules into the aggregation process of GNNs, enabling models to combine the generalization capabilities of neural networks with the interpretability of symbolic logic. 

Recently, causal inference has been introduced into KGC to address issues of data distribution imbalance and spurious associations. KGCF \cite{DBLP:conf/www/ChangCL23} is a typical example of this approach. This method significantly improves the performance of knowledge graph completion by considering the imbalanced distribution of relationships. Similarly, the CFLP \cite{DBLP:conf/icml/ZhaoLW0022} model generates counterfactual links and utilizes the causal relationship between the global graph structure and link existence to further optimize the link prediction results.

\subsection{Inference for Prediction}

In prediction-oriented knowledge graph reasoning methods, the key goal is to learn evolution patterns from the historical states of the temporal knowledge graph (TKG) to effectively predict future states, as shown in Figure \ref{fig:TKGC}. The prediction method based on GNNs explores the intrinsic topological relationships between entities and between entities and relations in the TKG. This enables the acquisition of high-quality embedding representations.

Graph convolution-based methods typically employ a combined framework of a graph structure encoder and a time encoder to generate entity representations. Specifically, it extracts topological features through a structural encoder and captures sequence changes with a temporal encoder. Early studies such as RE-NET \cite{jin2019recurrent} laid the foundation. Subsequent approaches such as DHE-TKG \cite{DBLP:conf/coling/LiuZML0Z24}, L$^2$TKG \cite{DBLP:conf/acl/ZhangXLWW23}, RPC \cite{DBLP:conf/sigir/0006MLLTWZL23}, and EvoKG \cite{DBLP:conf/wsdm/ParkLMCFD22} all followed this line of thinking, with the core idea being to combine relational graph convolutions (R-GCN) with recurrent neural networks (such as GRU). Among these, EvoKG specifically emphasizes the joint modeling of event timing and structural evolution, while RPC further enhances the representational consistency of entities across extended time spans by introducing cross-temporal similarity graphs. Given the characteristics of sparse historical data and recurring events, another research focus has shifted toward complex composition strategies. STDN \cite{DBLP:journals/tkdd/LiZYLLL24}, MTKGE \cite{DBLP:conf/www/ChenXS0D23}, TiRGN \cite{DBLP:conf/ijcai/LiS022}, HGLS \cite{DBLP:conf/www/ZhangX0WW23}, HiMatch \cite{li2022hismatch}, and TiPNN \cite{DBLP:journals/ai/DongWXNWZ24} all employ temporal sequence patterns or historical subgraphs as their compositional framework. These models overcome the limitations of single-time-step snapshots by extracting historical sequences or global temporal patterns to address the issue of insufficient short-term information. TANGO \cite{DBLP:conf/emnlp/HanDMGT21} takes this approach a step further by utilizing the NODE (Necessary Ordinary Differential Equation) framework to transform the discrete GCN update process into a continuous temporal flow model.

Graph attention-based reasoning methods achieve information aggregation and node embedding updates by assigning differentiated weights to neighboring nodes. To enhance inference efficiency and filter out noise, T-GAP \cite{DBLP:conf/kdd/JungJK21} introduces a subgraph sampling mechanism that combines with a time-aware attention mechanism to capture spatio-temporal correlations within local spaces. TPmod \cite{DBLP:journals/tkdd/BaiMZY21} and DA-Net \cite{DBLP:conf/cikm/LiuZC0X022} both approach the problem from the perspective of attention weights. The former measures the importance of historical events by defining a “propensity policy” between entities, while the latter employs distributed attention to resolve the “role overlap” issue caused by fixed vector summation in traditional models. Recent research has increasingly focused on capturing deeper interactions through multi-view architectures. TaReT \cite{DBLP:journals/ipm/MaLZWLLQ24}, EvoExplore \cite{DBLP:journals/kbs/ZhangLSS22}, and HGAT \cite{DBLP:journals/ijon/ShaoHL0T23} have all moved beyond single-graph structures to construct complex graphs incorporating co-occurrence facts, topological relationships, local-global views, or dual dimensions of temporal-semantic dimensions. This multidimensional modeling approach significantly enhances representation accuracy in complex scenarios. Furthermore, to balance modeling depth and breadth, LSGAT \cite{DBLP:conf/aaai/ChenYTCLZZ24} employs a hybrid architecture that leverages hierarchical R-GAT to capture short-term dependencies while integrating R-GCN to extract long-term global features. To address the challenge of event concurrency, LMS \cite{DBLP:journals/eswa/ZhangHMT24} and CRNet \cite{DBLP:conf/semweb/WangCZY22} explored semantic relationships among concurrent facts. LMS optimizes node representations by imposing temporal semantic constraints during attention computation, while CRNet innovatively utilizes candidate event graphs to simultaneously consider interactions between past and future concurrent events during inference.

\section{Knowledge Graph Applications}
In the field of knowledge graph applications, the core objective is to effectively integrate structured background knowledge into practical business scenarios, thereby enhancing semantic understanding and reasoning capabilities for downstream tasks. To address diverse business requirements, we have segmented this domain into major branches including Knowledge Graph Question Answering (KGQA), Recommendation Systems, Natural Language Processing (NLP),  Drug-Drug Interaction (DDI) prediction, and KG-augmented LLMs. As reflected in the summary presented in Table \ref{tab:kg-applications}, we analyze and distill these representative methods based on the two fundamental dimensions of graph construction strategies and GNN architectures.
\label{KGA}

\begin{figure}[!htbp]
    \centering
    \begin{minipage}[c]{0.49\linewidth}
        \vspace{0pt} 
        \begin{subfigure}[b]{\linewidth}            \includegraphics[width=0.98\linewidth]{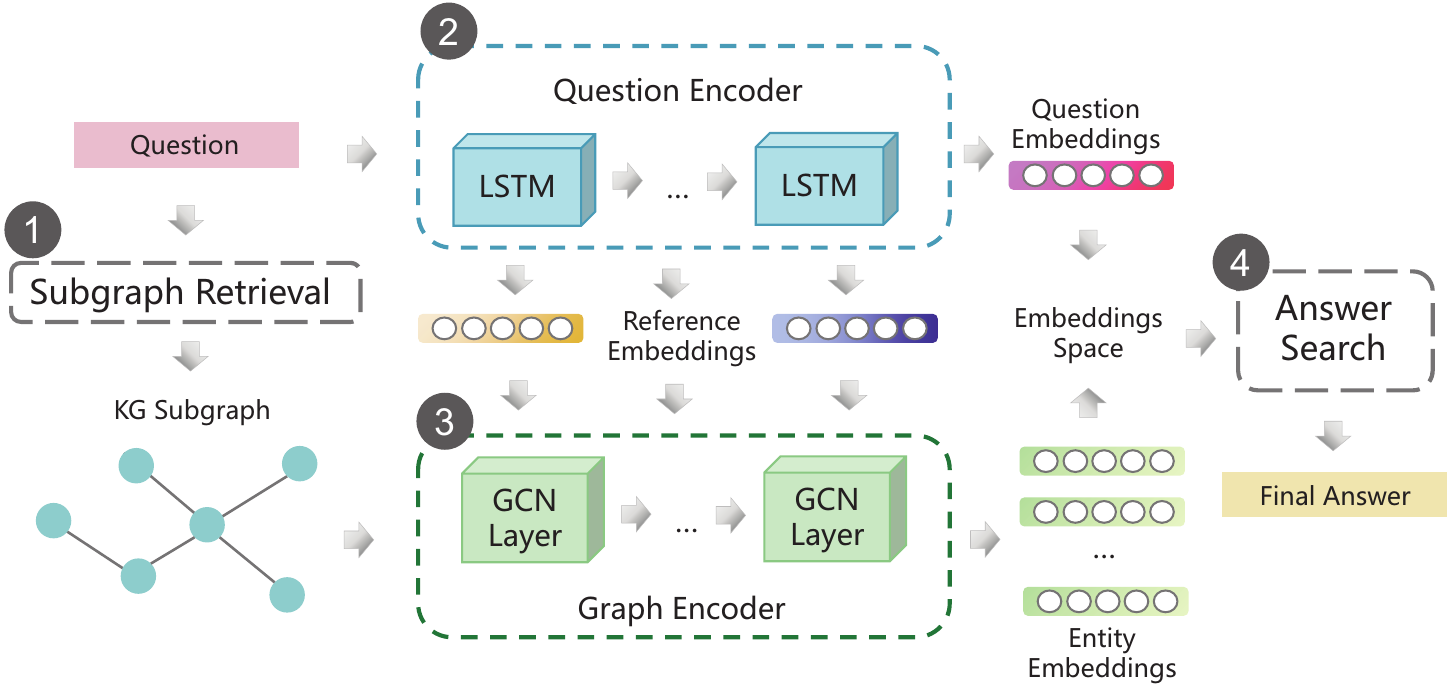}
            \caption{Question answering}
            \label{fig:KGQA}
        \end{subfigure}        
        \vspace{5pt} 
        \begin{subfigure}[b]{\linewidth}            \includegraphics[width=0.98\linewidth]{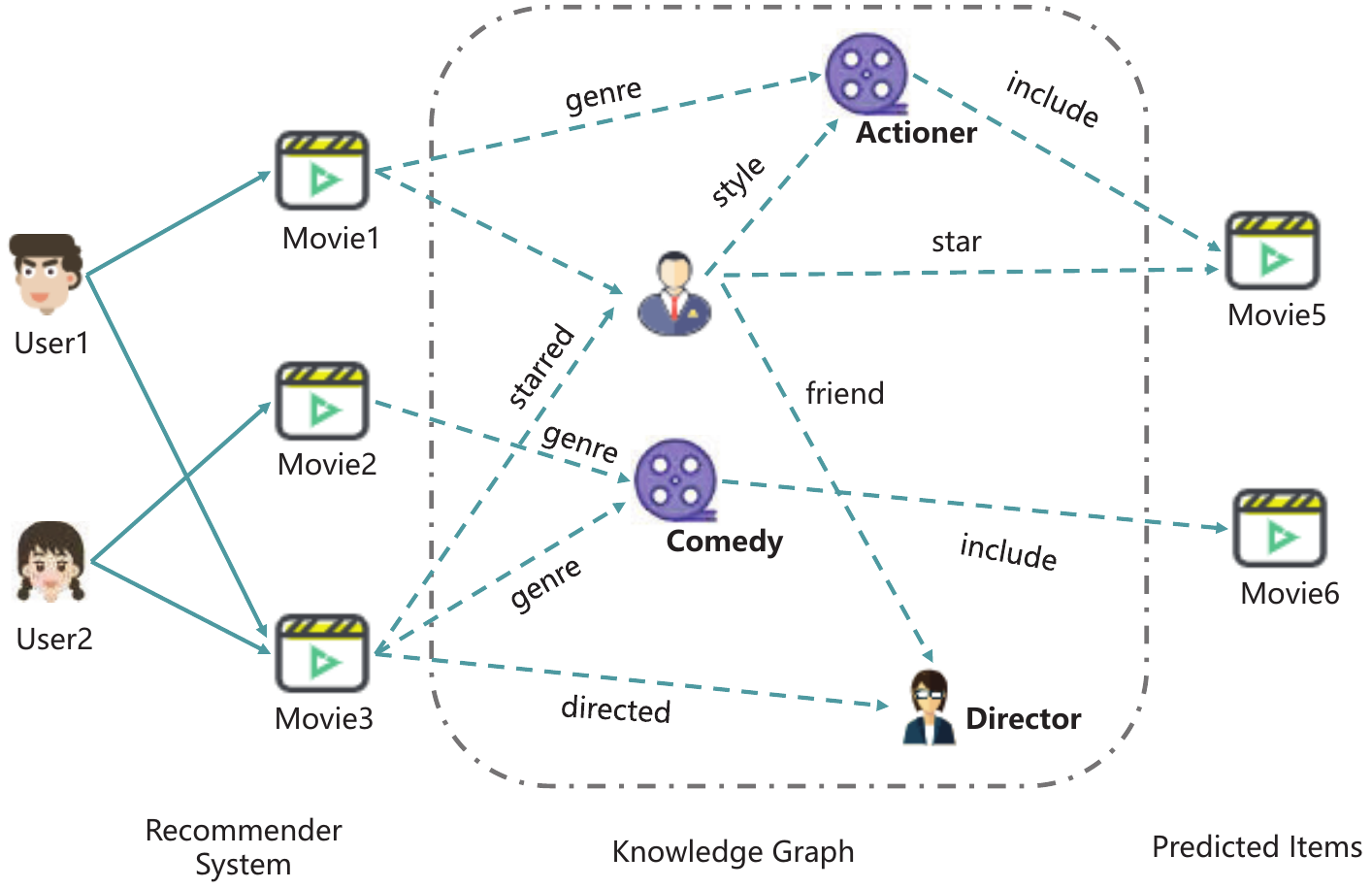}
            \caption{Recommender systems}
            \label{fig:recommender_sysytem}
        \end{subfigure}
    \end{minipage}
    \begin{minipage}[c]{0.49\linewidth}
        \vspace{0pt} 
        \begin{subfigure}[b]{\linewidth}            \includegraphics[width=0.98\linewidth]{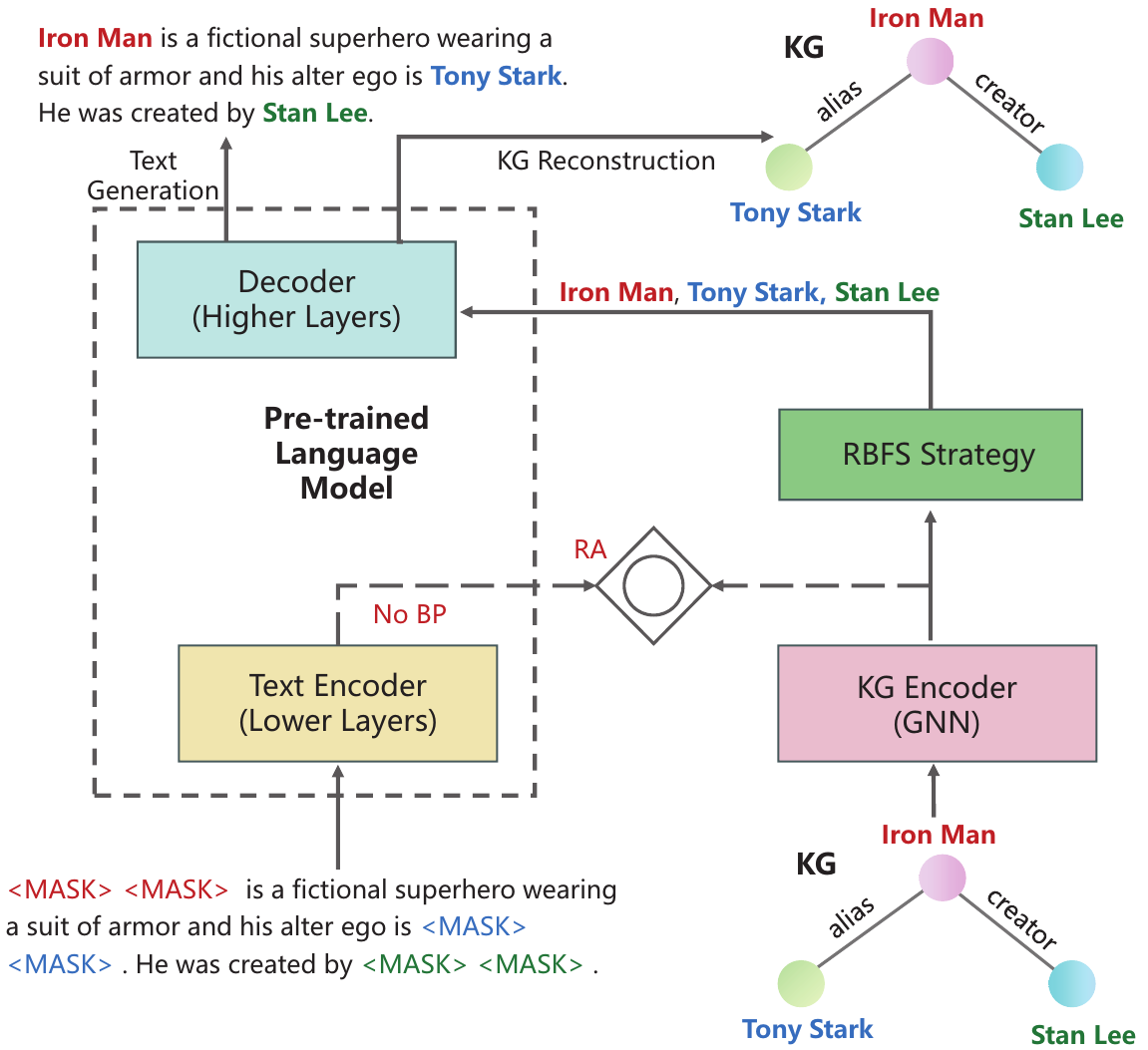}
            \caption{NLP-to-KG conversion}
            \label{fig:NLP}
        \end{subfigure}        
        \vspace{5pt} 
        \begin{subfigure}[b]{\linewidth}            \includegraphics[width=0.98\linewidth]{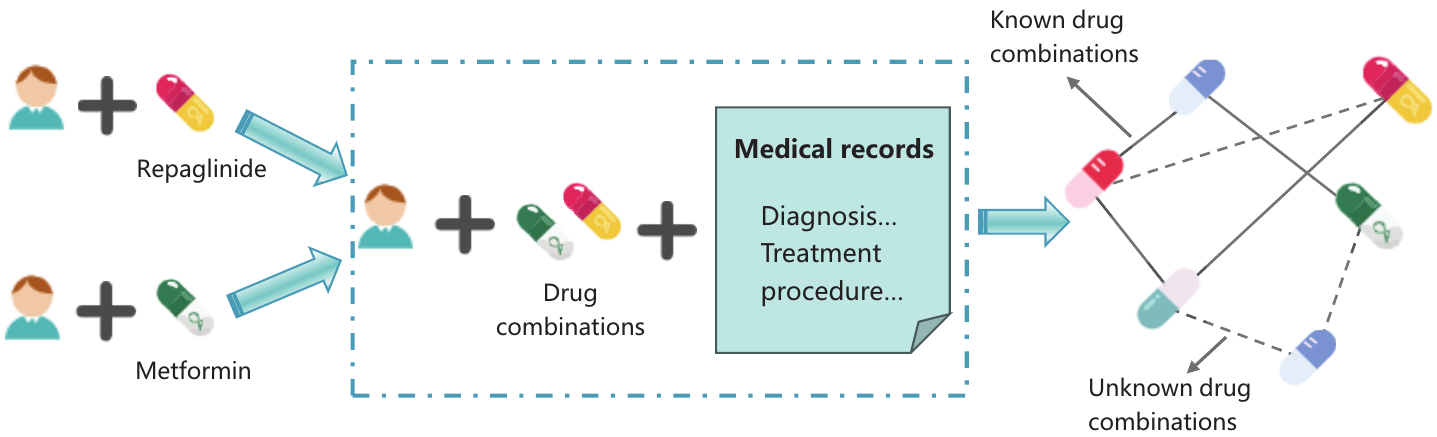}
            \caption{Drug interaction prediction}
            \label{fig:DDI}
        \end{subfigure}
    \end{minipage}
    \caption{Applications of knowledge graphs in AI systems}
    \label{fig:KG_Applications}
\end{figure}

\subsection{Knowledge Graph Question Answering}

Knowledge Graph Question Answering (KGQA) is a natural language processing technology based on knowledge graphs, aiming to extract relevant information from a knowledge graph and generate precise answers to user queries \cite{DBLP:conf/emnlp/JiangZZLW23}, as shown in Figure \ref{fig:KGQA}. A key challenge for KGQA systems lies in effectively handling both natural language queries and the heterogeneous information in the knowledge graph \cite{DBLP:conf/www/ZamiriQNZ024}. GNNs offer an effective solution to this problem. They can learn contextual relationships and entity representations within the knowledge graph, enhancing the system's reasoning capabilities through the use of the graph structure.
In KGQA systems, question understanding and parsing are usually the first steps. The core task in this phase is to extract semantic information from the natural language question, identify the question's intent and context, and lay the foundation for subsequent graph querying and reasoning. Both HamQA \cite{dong2023hierarchy} and JointLK \cite{DBLP:conf/naacl/SunSQZ22} focus on establishing a joint representation between questions and knowledge graphs. The former captures semantic relationships through hierarchical attention, while the latter achieves fusion and updating of question representations and knowledge graph nodes by tightly coupling language models with GNNs. ACENet \cite{DBLP:conf/emnlp/HaoXZ22} and DELFT \cite{DBLP:conf/www/ZhaoXQB20} further expand upon this foundation by introducing external common-sense knowledge and semantically dense queries, respectively, to address reasoning biases caused by insufficient semantic information in the problem.

In the graph query generation and reasoning phase, the focus of research is on retrieving relevant information from the knowledge graph and generating query or reasoning paths to support answering the question. Subgraph reasoning is a common reasoning method, aiming to focus on the key information related to a specific question by constructing subgraphs. QA-GNN \cite{DBLP:conf/naacl/YasunagaRBLL21} and SR \cite{DBLP:conf/acl/ZhangZY000C22} are representative approaches for subgraph reasoning. They confine reasoning to a local space by constructing specific subgraphs that incorporate the problem context. CBR-SUBG \cite{DBLP:conf/icml/DasGNTZHJM22} takes a different approach by leveraging structural similarities between subgraphs to handle queries requiring subgraph reasoning. To address multi-relationship and temporal information, QAGCN \cite{DBLP:conf/esws/WangRCB24} proposes a question-aware convolutional approach that enables single-step implicit inference within a unified embedding space. Exaqt \cite{jia2021complex}, meanwhile, is specifically optimized for temporal facts by enriching subgraph representations through the introduction of temporal features. Furthermore, Graph2Seq \cite{DBLP:journals/tnn/ChenWZ24} introduces a bidirectional gated GNN that optimizes the process of information propagation and processing. After completing the inference process, the system enters the candidate answer screening phase. The current trend involves employing multi-stage interactive mechanisms (such as ACENet's multi-stage answer selection interaction), where feedback signals guide the GNN to perform more efficient feature propagation and ensure the selection of optimal answers.

\subsection{Recommendation System}
Recommendation system is a pivotal tool for internet platforms to uncover potential user interests \cite{fan2021continuous, DBLP:conf/cikm/FanL00Y21, yang2021consisrec}. As shown in Figure \ref{fig:recommender_sysytem},it relies on user-item interaction data to extract collaborative signals, which are then used to recommend similar content to users \cite{DBLP:conf/aaai/XiaHXDZYPB21}. Many studies have proposed leveraging knowledge graphs as auxiliary information \cite{DBLP:conf/aaai/LiuW0PY21} to enhance the capabilities of recommendation systems. By establishing connections between users' historical preferences and recommended items, KGs not only improve recommendation relevance but also enhance the interpretability of the system\cite{DBLP:conf/www/WangZXLG19}. GNNs effectively aggregates KG-based item interactions through its information propagation mechanism, enhancing recommendation systems\cite{DBLP:journals/fgcs/ZhangWWRC23, DBLP:conf/cikm/LiCWZW19, DBLP:conf/aaai/HuangXXDXLBXLY21, DBLP:conf/cikm/FanLH0WXY23,DBLP:conf/sigir/FengHLLZO20}.

To extract key features from complex knowledge graphs, one class of research focuses on reducing noise through subgraph partitioning or sampling mechanisms. Both DSKReG \cite{DBLP:conf/cikm/WangLFSY21} and GraphSW \cite{DBLP:journals/corr/abs-1908-05611} focus on optimizing sampling strategies. The former employs Gumbel-Softmax to achieve differentiable sampling, while the latter avoids information constraints imposed by fixed neighborhoods through staged resampling. In contrast, SARC \cite{DBLP:conf/ictai/ZhangCX19} adopts a decomposition strategy, breaking down user-project-entity interactions into three bidirectional interaction graphs and fusing information through a gating mechanism. KLGCN \cite{DBLP:journals/eswa/WangLZW22} achieves significant parameter complexity reduction while preserving subgraph aggregation performance through simplified feature transformation and nonlinear activation.

Another category of research focuses on capturing semantic information and user personalization within knowledge graphs through attention mechanisms. Like KGAT \cite{DBLP:conf/kdd/Wang00LC19}, it explicitly models higher-order relationships in knowledge graphs through recursive propagation and attention weights. CGAT \cite{DBLP:journals/tkde/LiuYXMWZ23} and HAGERec \cite{DBLP:journals/kbs/YangD20} further incorporate personalized and hierarchical attention mechanisms to capture user-specific context and multi-level preferences within heterogeneous knowledge graphs, respectively. Additionally, KGIN \cite{DBLP:conf/www/WangHWYL0C21} models user intent as a combination of KG relationships, while DIKGNN \cite{DBLP:conf/cikm/TuQWZLZWZZ23} captures multi-intent representations for fund recommendations through attribute nodes. KGCN \cite{DBLP:conf/www/WangZXLG19} and KGNN-LS \cite{DBLP:conf/kdd/WangZZLZLW19} enhance the model's generalization ability for handling long-tail interests by employing user association weight sampling and label smoothing regularization, respectively. Traditional GNNs based on simple graphs struggle to directly capture complex connections beyond binary relationships. To address this, SDK \cite{DBLP:conf/cikm/LiuXLWCY23} introduces a hypergraph self-supervised learning mechanism, mitigating GNN over-smoothing by constructing user-project hypergraphs. HyperTenet \cite{DBLP:conf/icdm/VijaikumarHS21} builds a 3-uniform hypergraph, leveraging a self-attention-based hypergraph neural network to capture triadic relationships among users, projects, and lists.

\subsection{Natural Language Processing}
Natural Language Processing (NLP) focuses on enabling computers to understand, generate, process, and respond to natural language text or speech through algorithms and models\cite{priya2024computational}. Pre-trained language models (PLMs) leverage large-scale unlabeled corpora for self-supervised training \cite{hu2023survey}. They have achieved remarkable performance across various NLP tasks, as exemplified by models such as BERT \cite{devlin2019bert}, RoBERTa \cite{liu2019roberta}, XLNet \cite{yang2019xlnet}, and the GPT series \cite{radford2019language, brown2020language, achiam2023gpt}. Research has shown that PLMs can effectively capture linguistic patterns in text and generate high-quality context-aware representations. However, these models often struggle to grasp world knowledge about entities and relationships, which is crucial for language understanding. Integrating knowledge into PLMs can enhance their memory and reasoning capabilities \cite{liu2021kg,DBLP:conf/ijcnlp/SchneiderSVGSM22}. 

In the research of knowledge graph-augmented PLMs, GNNs are often used to model subgraphs extracted from knowledge graphs, thereby enhancing the model's reasoning capabilities in Figure \ref{fig:NLP}. These models improve the performance of natural language processing by integrating graph structural information and language representations. Specifically, GNNs are employed to model the extracted subgraphs, and the coupling of PLMs with GNN modules enables joint reasoning, boosting performance in commonsense reasoning tasks. For example, JointLK \cite{DBLP:conf/naacl/SunSQZ22} and GreaseLM \cite{DBLP:conf/iclr/0001BYRLML22} use GNNs to model the extracted knowledge graph and couple the language model with the GNN module to perform joint reasoning for commonsense reasoning. Similarly, ReasoningLM \cite{DBLP:conf/emnlp/JiangZZLW23} introduces a subgraph-aware self-attention mechanism, allowing PLMs to directly support structured reasoning over knowledge graphs, significantly improving performance in knowledge graph question answering tasks. Additionally, JAKET \cite{DBLP:conf/aaai/Yu0Y022} proposes a joint pre-training framework, where the synergy between the knowledge module and the language module combines entity and relationship embeddings from the knowledge graph with text representations, offering a new optimization path for semantic analysis. For domain-specific applications, GOProteinGNN \cite{DBLP:conf/cikm/KalifaSR25}, while focusing on protein representation learning, provides a method of enhancing model representation by integrating protein knowledge graph information, which also offers new insights for natural language understanding. In the field of natural language generation, ASGARD \cite{DBLP:conf/acl/HuangWW20} encodes knowledge graph triples through a graph encoder and constructs a model capable of capturing global interactions between entities and topic transitions for abstractive summarization. Another example is the work of Li et al. \cite{DBLP:conf/acl/LiTZWYW21}, where their model uses R-GCN to encode triples and aligns the encoded entity embeddings with the PLM, effectively addressing the knowledge graph-to-text generation problem in a few-shot setting. CogNLG \cite{DBLP:journals/es/LaiYFCWWC24} extracts dynamic knowledge from subgraphs to generate factually consistent text and visualizes the reasoning paths within the knowledge graph, demonstrating the logical process of text generation.

\subsection{Drug-Drug Interaction}
Drug-drug interaction (DDI) prediction has emerged as a significant research trend in intelligent healthcare in recent years. As shown in Figure \ref{fig:DDI}, the advantage of knowledge graphs in DDI prediction lies in their ability to integrate multi-source heterogeneous medical data, including drug chemical compositions, molecular structures, target information, and complex relationships with other biological entities such as genes, proteins, and diseases \cite{DBLP:conf/aaai/ZhangLNZFM24}. This rich entity and relational information provides comprehensive semantic support for the study of drug interaction mechanisms, making knowledge graph-based drug discovery one of the most reliable research paradigms today.

To fully leverage the heterogeneous properties of drugs, many studies have focused on the fusion of multi-perspective features. MKG-FENN \cite{DBLP:conf/aaai/WuSHC024} and MUFFIN \cite{chen2021muffin} both adopt multi-channel fusion strategies. The former constructs four types of graph structures encompassing chemical entities and molecular structures, utilizing GCN to extract high-order semantic information. The latter employs a dual-layer cross-fusion mechanism to balance drug molecular structures with KG semantic information. MSKG-DDI \cite{chen2024effective}, HetDDI \cite{li2023hetddi}, and DeepLGF \cite{ren2022biomedical} further advance this approach by integrating drug-receptor graphs or pre-trained heterogeneous models to explore complementarity among multi-modal representations. To address the limitation of fixed weight distribution in traditional GNNs, LaGAT \cite{DBLP:journals/bioinformatics/HongLJL22} introduced a link-aware graph attention mechanism to generate customized attention paths for different drug pairs. MK-GNN \cite{gao2023medical} adopts a patient-centric approach to extracting multidimensional features across distinct feature subspaces. Furthermore, GCLR \cite{an2024disease} represents the latest technological trend by introducing Graph Contrastive Learning into the DDI framework. By contrasting representations across different views, it significantly enhances the system's robustness against noisy data. 

Faced with the explosive growth of knowledge graphs, efficiently extracting key prediction paths has become a research priority. KGNN \cite{DBLP:conf/ijcai/LinQWMZ20} expands the model's receptive field to capture long-range dependencies by mining associations between drugs and their neighboring nodes. In contrast, SumGNN \cite{DBLP:journals/bioinformatics/YuHZGSX21} proposes a more streamlined subgraph extraction and generalization scheme. By generalizing k-hop local subgraphs, it significantly improves prediction accuracy for low-frequency DDI types.

\subsection{KG-augmented LLMs}
With the widespread adoption of LLMs, their inherent hallucination issues, knowledge lag, and lack of interpretability in complex logical reasoning have gradually become obstacles to their advancement into rigorous application domains. Meanwhile, KGs, as carriers of high-quality, structured facts, provide LLMs with reliable knowledge anchors. Based on the intervention stage and core objectives of knowledge graphs, existing research can be systematically categorized into three closely interrelated directions \cite{pan2024unifying}: knowledge graph-enhanced LLM pre-training, knowledge graph-enhanced LLM reasoning, and knowledge graph-enhanced LLM interpretability.

During the pre-training phase, the research focus lies in how to efficiently transform the structured facts stored in KGs into parameterized representations for LLMs. Representative methods such as ERNIE \cite{DBLP:journals/corr/abs-2107-02137} and DKPLM \cite{zhang2022dkplm}. To reduce the computational cost of full-parameter updates, KG-Adapter \cite{tian2024kg} and InfuserKI \cite{wang2150infuserki} provide effective solutions for lightweight encoding and anti-forgetting mechanisms, respectively. Pre-trained models often face the issue of knowledge obsolescence. GLAME \cite{zhang2024knowledge} proposes a novel model editing paradigm that leverages knowledge graphs to identify relevant knowledge affected by edits and synchronously updates the internal representations within LLMs.

During the reasoning phase, KG primarily functions as an external reliable knowledge base, guiding LLMs to perform logically rigorous reasoning through Retrieval-Augmented Generation (RAG) or structured prompts. Traditional RAG focuses on local fact retrieval, while GraphRAG \cite{edge2024local} extends it to global understanding by constructing entity knowledge graphs from documents. ChatKBQA \cite{luo2024chatkbqa} proposes an innovative “generation-retrieval” framework that first utilizes a fine-tuned LLM to generate logical forms, then employs unsupervised retrieval to precisely replace entities and relationships, thereby simplifying the process and enhancing performance. Additionally, K-RagRec \cite{wang2025knowledge} applies KG retrieval enhancement to recommendation systems to address issues such as LLM hallucinations and insufficient domain knowledge. Notably, WTS \cite{zhang2025way} proposes a dynamic closed-loop paradigm that not only leverages KGs to enhance LLMs but also utilizes domain knowledge generated by these models to enrich the KGs, achieving synergistic model specialization and knowledge base evolution.

KG's structured nature provides a natural interpretive tool and evaluation benchmark for the “black box” of LLM, helping mitigate hallucinations and enhance model credibility. Methods such as Mindmap \cite{wen2024mindmap}, RoG \cite{luo2024reasoning}, ChatRule \cite{luo2025chatrule}, and CoK \cite{wang2024boosting} enable LLMs to generate explicit reasoning chains based on KG entity relationships, thereby not only making the reasoning process transparent but also facilitating verification.  Frameworks such as KGR \cite{guan2024mitigating} and OKGQA \cite{DBLP:conf/acl/SuiHDH25} focus on fact-checking and correction after inference.

\section{Open issues and future directions}

\label{issues}

{\textbf{Multimodal fusion and cross-domain.} KGs in the real world are evolving toward multimodality. Recent research \cite{tang2024graphgpt} has begun leveraging the powerful semantic alignment capabilities of LLMs to bridge KGs with multimodal data. Although LLM demonstrates significant potential in aligning image pixels, audio waveforms, and structured triples, existing collaborative models often “flatten” graph structures into linear text, resulting in substantial loss of topological information \cite{DBLP:conf/iclr/Zhu0SLE24, girdhar2023imagebind}. To achieve optimal integration between LLMs and knowledge graphs, model architectures must go beyond simple concatenation. LLMs are increasingly driving the design of GNNs, requiring GNNs to serve as lightweight structural prompts generated via cross-modal projectors. To balance the efficiency of freezing the LLM with the topological accuracy of full fine-tuning, future architectures should combine adjustable GNN encoders with parameter-efficient fine-tuning on the LLM. Furthermore, the computational overhead associated with processing high-dimensional modal data remains substantial. Therefore, the core challenge for the present and future lies in enabling models to accurately perceive fine-grained graph structures and multimodal associations without requiring costly pre-training.}

{\textbf{Real-time updates of dynamic knowledge graphs.}
Knowledge graphs in the real world exhibit high heterogeneity and dynamic evolution. While traditional research has employed temporal GNNs, memory mechanisms, and incremental learning to capture graph structure evolution, a pressing challenge remains: how to efficiently maintain historical information and reduce computational overhead during real-time updates while preserving model expressiveness. Recently, the introduction of LLMs has opened new avenues in this field. Researchers are beginning to explore leveraging LLMs to extract temporal logic rules \cite{wang2024large} and integrate language model semantics with graph topology structures for knowledge editing \cite{lu2025knowledge, ong2025dynamic}. Despite this, LLMs still face hallucination risks in temporal reasoning within specific domains. The core challenge for the future lies in leveraging the explicit structural constraints of GNNs to calibrate the reasoning process of LLMs, while achieving low-latency, high-consistency collaborative knowledge editing and conflict arbitration in large-scale dynamic environments.}

{\textbf{Low-resource and few-shot learning.} In many practical applications, the construction and reasoning of knowledge graphs rely on high-quality labeled data. However, the cost of obtaining labeled data is high, especially in specialized domains or minority languages \cite{lan2025nlp, li2024flexkbqa}. Recent advances have attempted to utilize LLMs as “data augmenters” or “logical reasoners” to assist GNNs in performing structural predictions under small-sample conditions \cite{ye2022ontology, wang2024llm, zhang2024gail}. Although LLMs possess rich prior knowledge, their issue of hallucination continues to undermine their reliability as a credible knowledge source in specialized domains. Future research should move beyond simple data augmentation based on large language models and instead explore fine-grained logical alignment frameworks. Specifically, emphasis should be placed on leveraging the explicit symbolic logic inherent in knowledge graphs to impose real-time path constraints and factual verification on the generative processes of LLMs. This approach is essential to ensure the consistency of collaborative systems when addressing long-chain reasoning tasks.}

{\textbf{Scalability and computational efficiency.} As the scale of KGs continues to expand, their representation learning and various downstream tasks face increasingly severe challenges in terms of computational efficiency and scalability. Although fused LLMs can enhance a model's perception of graph structures and improve reasoning performance \cite{xiao2024fuselinker,qiao2024graphllm}, even parameter-efficient fine-tuning methods struggle to avoid high resource consumption \cite{tian2024kg}. Furthermore, while techniques such as sparse GNNs, distributed computing, and graph sampling for large-scale graphs can reduce computational complexity, they often come at the cost of information loss, gradient instability, or degraded performance. To effectively design KG-LLM approaches capable of handling large-scale datasets without degrading performance, future research must prioritize scalable architectural co-designs. For example, by using a decoupled pipeline, computation-intensive graph structure encoding can be handled offline. Alternatively, before entering the LLM’s context window, the most relevant subgraphs can be precisely extracted and the graph tokens compressed to reduce the computational load~\cite{liu2025filter}. 
Furthermore, a highly promising direction is the exploration of Graph Foundation Models (GFMs)\cite{shi2024graph}. Inspired by the success of LLMs, GFMs aim to capture universal graph patterns through pre-training on diverse large-scale graphs\cite{DBLP:journals/pami/LiuYLCLZBFSYS25,yang2025benchmarking,DBLP:journals/tkde/ZhaoSLZLZ25,shi2024lecture}. This paradigm not only reduces the retraining costs for specific graphs via transfer learning but also facilitates cross-task unification. By utilizing a unified interface, GFMs can dismantle barriers between fragmented downstream tasks, thereby achieving true scalability across both data magnitude and task diversity. Future research should shift its focus from learning representations for specific graph structures toward developing pre-training paradigms that possess general graph pattern recognition capabilities and support efficient parameter fine-tuning. This approach will enable true scalability across varying data sizes and diverse tasks.}

{\textbf{Industrial Deployment and Real-World Challenges.} 
Knowledge Graphs have become the critical infrastructure underpinning modern industrial intelligence. From enhancing Google's search semantic understanding \cite{noy2019industry} to optimizing e-commerce recommendations at Amazon (COSMO \cite{yu2024cosmo}, KGLA \cite{guo2024knowledge}) and Alibaba (AliCG \cite{zhang2021alicg}), and further enabling precision medicine for rare diseases at LMU Klinikum \cite{suwer2025privacy}, KG has demonstrated formidable knowledge representation capabilities. Beyond initial paradigms, companies are actively deploying KG-LLM architectures. BenevolentAI leverages KGs to enhance LLMs and accelerate drug discovery~\cite{xu2025survey}. GraphRAG~\cite{edge2024local} has already established itself as a standard architecture in enterprise-level search. Furthermore, this convergence is extending into the metaverse. As a key enabling technology, KGs provide the underlying semantics for modeling virtual worlds~\cite{sun2025mskd, DBLP:conf/semweb/JaiminiZB22}. Utilizing the structural constraints of KGs to anchor LLM generation ensures logical consistency within digital environments and interoperability among virtual assets. However, in complex scenarios such as industrial standard parsing and clinical guideline application, KG still faces challenges including the difficulty of processing unstructured text and stringent compliance requirements. Although the ontological framework proposed by Jiin Park et al. \cite{park2025ontology} and OpenAI’s "KG+LLM" \cite{openai2025temporalagents} integration practices have offered initial paradigms to address the aforementioned bottlenecks, developing native interpretable reasoning architectures to overcome the limitations of existing plug-in-based specification checks remains a key research focus when handling complex industrial standards.}

\section{Conclusion}
\label{conclusion}
In recent years, GNNs have demonstrated significant advantages in the field of knowledge graphs due to their powerful graph-structured modeling capabilities. By enhancing the representation quality of entities and relations, GNNs have substantially improved the ability of knowledge graphs in data representation and reasoning. 
This survey first introduces the fundamental definitions of knowledge graphs and GNNs, covering the formal definition, types, and construction process. From the perspectives of knowledge graph techniques and GNN methods, we proposes a novel two-level taxonomy framework. Specifically, we categorize the KG techniques pipeline into four key phases: construction, embedding, reasoning, and applications. 
For each phase, we systematically review how GNN techniques are integrated to address various tasks. Specifically, we conduct a systematic review and profound synthesis of existing methodologies across two pivotal dimensions, namely graph construction strategies and GNN architectures. Furthermore, we discusses the advantages and limitations of representative models. Finally, we highlight open challenges and outline promising future research directions. 
The overarching goal of this survey is to provide methodological guidance and technical insights for researchers, thereby fostering innovation at the intersection of GNNs and knowledge graphs.

\begin{acks}
We thank the partial support funded by Basic Research Program of Jiangsu (No.BK20251653), the Fundamental Research Funds for the Central Universities (No.2025QN1023), National Natural Science Foundation of China (No.61876186), and the National Natural Science Foundation of China Joint Fund Key Project (No.U25B20138).

\end{acks}

\bibliography{Manuscript}


\appendix
\section{Appendices}
A list of tables: Tables 1–5 compare the models from each section. GNNs in knowledge graphs now cover a full spectrum, from information extraction and knowledge fusion to logical reasoning and multi-scenario applications.
\begin{table*}[htbp]
\centering
\caption{Summary of GNN-based models for Knowledge Extraction.}
\label{tab:Knowledge Extraction}
\small 
\renewcommand{\arraystretch}{1.1} 
\begin{tabularx}{\linewidth}{@{} c l l c Y @{}}
\toprule
\textbf{Task} & \textbf{Type} & \textbf{Model} & \textbf{GNN} & \textbf{Graph Construction} \\ 
\midrule
\multirow{13}{*}{NER} 
    & \multirow{4}{*}{\makecell{Text-Based\\Modeling}} 
    & M-DGNN \cite{gu2024enhanced} & GCN & Co-occurrence Graph, Dependency Syntax Graph \\
    & & GNNER \cite{zaratiana2022gnner} & GCN+GAT & Text span symbol graph \\
    & & RWGNN \cite{chen2023randomly} & MPNN & Neural network structure graph \\
    & & DGNN \cite{liu2022learning} & GAT & Dependency Syntax Graph \\
\cmidrule(lr){2-5}
    & \multirow{5}{*}{\makecell{Integrating\\external knowledge}} 
    & LGN \cite{gui2019lexicon} & GAT & Character - word - sentence three-level graph \\
    & & MCGAT \cite{zhao2021multi} & GAT & Relationship type graph, word frequency weight graph, position embedding graph \\
    & & PGAT \cite{wang2022polymorphic} & GAT & Character-word polymorphism graph \\
    & & Kana \cite{nie2021knowledge} & GAT & Text-entity graph \\
    & & BAC-GNN-CRF \cite{xu2024semantic} & GAT & Character - word - chapter graph \\
\cmidrule(lr){2-5}
    & \multirow{4}{*}{\makecell{Complex Task\\Scenarios}} 
    & Trigger-GNN \cite{sui2022trigger} & GCN/GGNN & Vocabulary directed graph \\
    & & HGFNER \cite{zhang2024chinese} & GCN & Spatial structure Graph, Dependency Syntax Graph \\
    & & RGCN \cite{zhao2022learning} & GCN & Text-image graph, image-image graph \\
    & & MOUSING \cite{lu2024few} & GCN & Text-image graph \\

\midrule
\multirow{14}{*}{RE} 
    & \multirow{5}{*}{Sentence-level} 
    & AGGCN \cite{DBLP:conf/acl/GuoZL19} & GAT & Dependency Syntax Graph \\
    & & LSTAGCN \cite{sun2020relation} & GCN & Weighted Dependency Syntax Graph \\
    & & DAGCN \cite{li2021improve} & GAT & Dependency Syntax Graph \\
    & & A-GCN \cite{DBLP:conf/acl/TianCSW20} & GAT & Dependency Syntax Graph(Global + Local) \\
    & & C-GCN \cite{DBLP:conf/emnlp/Zhang0M18} & GCN & Dependency Syntax Graph \\
\cmidrule(lr){2-5}
    & \multirow{6}{*}{Document-level} 
    & DUALGRAPH \cite{DBLP:journals/tnn/LiFC24} & GCN & Entity Graph, Entity distribution Graph \\
    & & RECON \cite{DBLP:conf/www/BastosN0MSHK21} & GAT & Dependency Syntax Graph, Neighborhood subgraph \\
    & & HeterGSAN \cite{xu2021document} & GAT & Mention-Entity-Sentence graph \\
    & & GAIN \cite{DBLP:conf/emnlp/ZengXCL20} & GCN & Mention graph, Entity graph \\
    & & EoG \cite{DBLP:conf/emnlp/ChristopoulouMA19} & Path-based GNN & Mention-Entity-Sentence graph \\
    & & GLRE \cite{DBLP:conf/emnlp/WangHCS20} & GCN & Mention-Entity-Sentence graph \\
\cmidrule(lr){2-5}
    & \multirow{3}{*}{Adaptive Graph} 
    & GDPNet \cite{xue2021gdpnet} & GCN & Gaussian Graph \\
    & & LSR \cite{DBLP:conf/acl/NanGSL20} & Dynamic GNNs & Latent Dynamic Graph \\
    & & GP-GNN \cite{DBLP:conf/acl/ZhuLLFCS19} & Dynamic GNNs & Entity Graph \\

\midrule
\multirow{10}{*}{\makecell{Joint\\extraction}} 
    & \multirow{4}{*}{Syntax-driven} 
    & GraphRel \cite{fu2019graphrel} & GCN & Dependency Syntax Graph, Relation weight graph \\
    & & CPJE \cite{wang2022conditional} & GCN & Dependency Syntax Graph \\
    & & TAG \cite{zhang2023novel} & GCN & mention graph \\
    & & DGAT \cite{zhao2024joint} & GAT & word graph \\
\cmidrule(lr){2-5}
    & \multirow{3}{*}{\makecell{Novel graph\\structure}} 
    & ESEI \cite{li2023joint} & GAT & Span graph, Type graph \\
    & & RIFRE \cite{zhao2021representation} & GAT & entity-relation graph \\
    & & Bio-RIFRE \cite{esmail2022chemical} & GAT & entity-relation graph \\
\cmidrule(lr){2-5}
    & \multirow{3}{*}{\makecell{Complex semantic\\scenarios}} 
    & HGERE \cite{DBLP:conf/emnlp/0001YLT23} & HyperGNN & entity-relation graph \\
    & & MVGNAS \cite{al2022multi} & GCN+GAT & Span graph \\
    & & BanglaAutoKG \cite{DBLP:conf/coling/WasiRIC24} & GCN+GAT & entity graph \\

\bottomrule
\end{tabularx}
\end{table*}

\begin{table*}[htbp]
  \caption{Summary of GNN-based models for Knowledge Fusion.}
  \label{tab:Knowledge Fusion}
  \small %
  \centering
  \renewcommand{\arraystretch}{1.3} 
  \setlength{\tabcolsep}{2pt}        
  \begin{tabularx}{\textwidth}{@{} l l l c Y @{}}
    \toprule
    \textbf{Task} & \textbf{Type} & \textbf{Model} & \textbf{GNN} & \textbf{Graph Construction} \\ 
    \midrule
    \multirow{10}{*}{\makecell[l]{Entity\\Alignment}} 
        & \multirow{4}{*}{Global Semantics} 
        & GNED \cite{DBLP:journals/kbs/HuDSSL20} & GCN & Entity-word graph \\
        & & DGCN \cite{DBLP:conf/www/WuZMGSH20} & Dynamic GNNs & mention-entity graph \\
        & & SGEL \cite{DBLP:conf/www/FangCLZLW20} & GAT & mention-entity graph \\
        & & CoGCN \cite{DBLP:journals/es/JiaCSD21} & GCN & entity graph \\
    \cmidrule(lr){2-5}
        & \multirow{2}{*}{Complex Topology} 
        & EL-Graph \cite{DBLP:conf/nlpcc/WangXK23} & GAT & mention-entity graph \\
        & & HEGEL \cite{DBLP:journals/dint/ChenWFZ22} & GCN & mention-entity-keyword graph \\
    \cmidrule(lr){2-5}
        & \multirow{3}{*}{Domain-specific} 
        & GNN-RE \cite{yin2023two} & GAT & BIM ontology graph, Dependency Syntax Graph \\
        & & ED-GNN \cite{DBLP:conf/sigmod/Vretinaris0EQO21} & GCN+GAT & Knowledge graph, query graph \\
        & & MMR-FEK \cite{DBLP:conf/coling/LuZPGW24} & GCN & mention-entity graph \\

    \midrule
    \multirow{22}{*}{\makecell[l]{Entity\\Linking}} 
        & \multirow{5}{*}{Relational Modeling} 
        & RNM \cite{zhu2021relation} & GCN & entity graph \\
        & & MRAEA \cite{mao2020mraea} & GAT & entity graph \\
        & & HGCN \cite{DBLP:conf/emnlp/WuLFWZ19} & GCN & entity graph \\
        & & VR-GCN \cite{ye2019vectorized} & GCN & entity graph \\
        & & RDGCN \cite{DBLP:conf/ijcai/WuLF0Y019} & GCN+GAT & entity graph, relation graph \\
    \cmidrule(lr){2-5}
        & \multirow{13}{*}{Structural Heterogeneity} 
        & EPEA \cite{wang2020knowledge} & GAT & Pair-wise Connectivity Graph \\
        & & AliNet \cite{sun2020knowledge} & GCN+GAT & entity graph \\
        & & NMN \cite{DBLP:conf/acl/WuLFWZ20} & GCN+GAT & Sampling Subgraph \\
        & & AttrGNN \cite{DBLP:conf/emnlp/LiuCPLC20} & GAT & Name/Literal/Digital/Structure graph \\
        & & ACK-MMEA \cite{li2023attribute} & GCN & Entity graph, text graph, image graph \\
        & & EVA \cite{liu2021visual} & GCN & entity graph \\
        & & TEA \cite{liu2023tea} & GCN+GAT & entity graph, time graph \\
        & & TEA-GNN \cite{DBLP:journals/corr/abs-2203-02150} & GAT & Temporal Quartet graph \\
        & & MuGNN \cite{DBLP:conf/acl/CaoLLLLC19} & GAT & entity graph \\
        & & SSP \cite{wong2020global} & GCN & entity graph \\
        & & HMAN \cite{DBLP:conf/emnlp/YangZSLLS19} & GCN & Entity-Attributes Graph \\
        & & STEA \cite{DBLP:conf/coling/CaiMMYZL22} & GCN & Temporal Quartet graph \\
        & & DualMatch \cite{liu2023unsupervised} & GAT & entity-time/entity-relation-time graph \\
    \cmidrule(lr){2-5}
        & \multirow{4}{*}{Complex Scenarios} 
        & DINGAL \cite{yan2021dynamic} & GCN & entity graph \\
        & & ActiveEA \cite{DBLP:conf/emnlp/LiuSZHZ21} & GCN & entity graph \\
        & & Lambda \cite{yinlambda} & GAT & Weighted Entity Graph \\
        & & LargeGNN \cite{xin2022large} & GAT & Sampling Subgraph \\

    \bottomrule
  \end{tabularx}
\end{table*}

\begin{table*}[htbp]
  \caption{Summary of GNN-based models for Knowledge Graph Embedding.}
  \label{tab:KGE}
  \small 
  \centering
  \renewcommand{\arraystretch}{1.3} 
  \setlength{\tabcolsep}{4pt}    
  \begin{tabularx}{\textwidth}{@{} l l l c Y @{}}
    \toprule
    \textbf{KG} & \textbf{Type} & \textbf{Model} & \textbf{GNN} & \textbf{Graph Construction} \\ 
    \midrule

    \multirow{7}{*}{\makecell[l]{Static\\KG}} 
        & \multirow{4}{*}{Novel GNN structure} 
        & R-GCN \cite{schlichtkrull2018modeling} & GCN & entity graph \\
        & & M-GNN \cite{wang2019robust} & GCN & entity graph \\
        & & MSeaKG \cite{di2023message} & GCN & entity graph, hypergraph \\
        & & SE-GNN \cite{li2022does} & GAT & entity graph \\
    \cmidrule(lr){2-5}
        & \multirow{3}{*}{Parameter Optimization} 
        & EARL \cite{chen2023entity} & GCN & entity graph \\
        & & decentRL \cite{guo2024distributed} & GAT & entity graph \\
        & & HKGN \cite{liu2022heterogeneous} & HGNN & entity graph \\

    \midrule
    \multirow{12}{*}{\makecell[l]{Dynamic\\KG}} 
        & \multirow{5}{*}{Inductive Methods} 
        & GraIL \cite{teru2020inductive} & GAT & Sampling Subgraph \\
        & & SNRI \cite{DBLP:conf/ijcai/XuZHCY22} & GAT & Sampling Subgraph \\
        & & InGram \cite{DBLP:conf/icml/LeeCW23} & GAT & entity graph, relation graph \\
        & & CoMPILE \cite{DBLP:conf/aaai/MaiZY021} & GAT & Sampling Subgraph \\
        & & DEKG-ILP \cite{DBLP:conf/icde/ZhangWYZCZ23} & GAT & Sampling Subgraph \\
    \cmidrule(lr){2-5}
        & \multirow{4}{*}{Enhanced Temporal Modeling} 
        & RE-GCN \cite{DBLP:conf/sigir/LiJLGGSWC21} & GCN & Temporal Sequence Pattern Graph \\
        & & GTRL \cite{DBLP:journals/tkde/TangC24} & GCN & entity graph, entity group graph \\
        & & TeMP \cite{DBLP:conf/emnlp/WuCCH20} & GCN & Temporal Sequence Pattern Graph \\
        & & DACHA \cite{DBLP:journals/tkdd/ChenTCQLZ22} & GCN & entity graph, relation graph \\
    \cmidrule(lr){2-5}
        & \multirow{3}{*}{Time-aware Attention} 
        & TARGAT \cite{DBLP:journals/taslp/XieZ0ZH23} & GAT & Temporal Sequence Pattern Graph \\
        & & xERTE \cite{DBLP:conf/iclr/HanCMT21} & GAT & Sampling Subgraph \\
        & & LogCL \cite{DBLP:conf/icde/ChenWWZCLL24} & GAT & local graph, global graph \\

    \midrule
    \multirow{7}{*}{\makecell[l]{Complex\\KG}} 
        & \multirow{5}{*}{Hyper-Relational KG} 
        & HeteHG-VAE \cite{DBLP:journals/pami/FanZWLZGD22} & HGNN & hypergraph, Dual Hypergraph \\
        & & H2GNN \cite{DBLP:journals/corr/abs-2412-12158} & HGNN & hypergraph \\
        & & DHGE \cite{DBLP:conf/aaai/LuoETZYW23} & Gform+HGNN & ontology graph, hypergraph \\
        & & STARE \cite{DBLP:conf/emnlp/GalkinTMUL20} & GCN & Hyper-Relational Graph \\
        & & QUAD \cite{DBLP:conf/asunam/ShomerJL0L23} & GCN & entity graph, Modifier Graph \\
    \cmidrule(lr){2-5}
        & \multirow{2}{*}{Multimodal KG} 
        & CMGNN \cite{DBLP:journals/tkde/FangZHWX23} & GAT & first-order subgraph, higher-order subgraph \\
        & & HRGAT \cite{DBLP:journals/tomccap/LiangZZ023} & GAT & Hyper-node graph \\

    \bottomrule
  \end{tabularx}
\end{table*}

\begin{table*}[htbp]
  \caption{Summary of GNN-based models for Knowledge Reasoning.}
  \label{tab:Knowledge Reasoning}
  \small %
  \centering
  \renewcommand{\arraystretch}{1.15} 
  \begin{tabularx}{\textwidth}{@{} l l l c Y @{}}
    \toprule
    \textbf{Task} & \textbf{Type} & \textbf{Model} & \textbf{GNN} & \textbf{Graph Construction} \\ 
    \midrule

    \multirow{18}{*}{\makecell[l]{Inference\\for\\Completion}} 
        & \multirow{6}{*}{\makecell[l]{Novel Graph\\Construction}} 
        & WGE \cite{DBLP:conf/esws/TongNPN23} & GCN & entity-relation graph \\
        & & TAGNet \cite{DBLP:conf/emnlp/Shomer0LWAT23} & GCN & Dynamic Graph \\
        & & INDIGO \cite{DBLP:conf/nips/LiuGHK21} & GCN & Entity Pair graph \\
        & & M2GNN \cite{DBLP:conf/www/WangWSWNAXYC21} & GAT & Multi-relationship graph \\
        & & LR-GCN \cite{DBLP:journals/fcsc/HeLCWZQ25} & GCN & entity graph, Virtual Edge \\
        & & NoGE \cite{DBLP:conf/wsdm/NguyenTPN22} & GCN & weight entity graph \\
    \cmidrule(lr){2-5}
        & \multirow{6}{*}{Enhanced Attention} 
        & RGHAT \cite{DBLP:conf/aaai/ZhangZZ0XH20} & GAT & entity graph \\
        & & DisenKGAT \cite{DBLP:conf/cikm/WuSCCLZW021} & GAT & entity graph \\
        & & GATH \cite{DBLP:journals/tkdd/WeiSY24} & GAT & entity graph \\
        & & MRGAT \cite{DBLP:journals/nn/DaiWZLC22} & GAT & entity graph \\
        & & DRR-GAT \cite{DBLP:journals/eswa/ZhangZGPNW23} & GAT & entity graph \\
        & & KRACL \cite{DBLP:conf/www/TanCFZZLL23} & GAT & entity graph \\
    \cmidrule(lr){2-5}
        & \multirow{4}{*}{Multi-source Fusion} 
        & LSMGA \cite{DBLP:conf/acl/TangZZZ23} & GAT & entity graph \\
        & & AR-KGAT \cite{DBLP:journals/kbs/ZhangHT22} & GAT & entity graph, Multi-jump auxiliary edge \\
        & & DMoG \cite{DBLP:conf/coling/SongHZG00022} & GCN+GAT & entity graph, Ontology graph, Text graph \\
        & & D-GNN \cite{wu2023graph} & GAT & entity graph, Ontology graph, Text graph \\
    \cmidrule(lr){2-5}
        & \multirow{2}{*}{Causal Inference} 
        & KGCF \cite{DBLP:conf/www/ChangCL23} & GCN & entity graph, Counterfactual graph \\
        & & CFLP \cite{DBLP:conf/icml/ZhaoLW0022} & GCN & entity graph, Counterfactual graph \\

    \midrule

    \multirow{21}{*}{\makecell[l]{Inference\\for\\Prediction}} 
        & \multirow{12}{*}{GCN-Based} 
        & RE-NET \cite{jin2019recurrent} & GCN+GAT & Temporal dynamic graph \\
        & & DHE-TKG \cite{DBLP:conf/coling/LiuZML0Z24} & GCN+HyperGNN & Temporal dynamic graph, Hypergraph \\
        & & L2TKG \cite{DBLP:conf/acl/ZhangXLWW23} & GCN+GAT & Temporal dynamic graph \\
        & & EvoKG \cite{DBLP:conf/wsdm/ParkLMCFD22} & GCN & Temporal Sequence Pattern Graph \\
        & & RPC \cite{DBLP:conf/sigir/0006MLLTWZL23} & GCN & \makecell[c]{entity graph, Relationship-related graph, \\ Cross-temporal similarity graph} \\
        & & STDN \cite{DBLP:journals/tkdd/LiZYLLL24} & GCN & temporal graph \\
        & & MTKGE \cite{DBLP:conf/www/ChenXS0D23} & GCN & \makecell[c]{Relative Position Pattern Graph, \\ Temporal Sequence Pattern Graph} \\
        & & TiRGN \cite{DBLP:conf/ijcai/LiS022} & GCN & Temporal Sequence Pattern Graph \\
        & & HiSMatch \cite{DBLP:conf/emnlp/LiHGJP0L0GC22} & GCN & Temporal Sequence Pattern Graph \\
        & & TANGO \cite{DBLP:conf/emnlp/HanDMGT21} & GCN & Multi-Relational Temporal Graphs \\
        & & HGLS \cite{DBLP:conf/www/ZhangX0WW23} & GCN+GAT & Temporal Sequence Pattern Graph \\
        & & TiPNN \cite{DBLP:journals/ai/DongWXNWZ24} & GCN & Temporal Sequence Pattern Graph \\
    \cmidrule(lr){2-5}
        & \multirow{9}{*}{GAT-Based} 
        & TPmod \cite{DBLP:journals/tkdd/BaiMZY21} & GAT & History Subgraph \\
        & & T-GAP \cite{DBLP:conf/kdd/JungJK21} & GAT & Sampling Subgraph \\
        & & TaReT \cite{DBLP:journals/ipm/MaLZWLLQ24} & GAT & \makecell[c]{Co-occurrence Fact Graph, Topological \\ Relationship Graph, Temporal Sequence Pattern} \\
        & & DA-Net \cite{DBLP:conf/cikm/LiuZC0X022} & GAT & Temporal Sequence Pattern Graph \\
        & & EvoExplore \cite{DBLP:journals/kbs/ZhangLSS22} & GAT & Temporal Sequence Pattern Graph \\
        & & LSGAT \cite{DBLP:conf/aaai/ChenYTCLZZ24} & GCN+GAT & Temporal Sequence Pattern Graph \\
        & & HGAT \cite{DBLP:journals/ijon/ShaoHL0T23} & GAT & Temporal Sequence Pattern Graph \\
        & & LMS \cite{DBLP:journals/eswa/ZhangHMT24} & GCN+GAT & Union Graph, Temporal Graph \\
        & & CRNet \cite{DBLP:conf/semweb/WangCZY22} & GAT & Temporal Graph, Candidate Event Graph \\

    \bottomrule
  \end{tabularx}
\end{table*}

\begin{table*}[htbp]
  \caption{Summary of GNN-based models for Knowledge Graph Applications.}
  \label{tab:kg-applications}
  \small %
  \centering
  \renewcommand{\arraystretch}{1.2} 
  \begin{tabularx}{\textwidth}{@{} l l c Y @{}}
    \toprule
    \textbf{Task} & \textbf{Model} & \textbf{GNN} & \textbf{Graph Construction} \\ 
    \midrule

    \multirow{10}{*}{\makecell[l]{Knowledge Graph\\Question\\Answering}} 
        & HamQA \cite{dong2023hierarchy} & GAT & question-Answer subgraph \\
        & JointLK \cite{DBLP:conf/naacl/SunSQZ22} & GAT & question-Answer subgraph \\
        & ACENet \cite{hao2022acenet} & GAT & Question-Answer-Retrieved entity subgraph \\
        & QAGCN \cite{DBLP:conf/esws/WangRCB24}  & GCN & Question subgraph \\
        & QA-GNN \cite{DBLP:conf/naacl/YasunagaRBLL21} & GAT & QA-KG Graph \\
        & CBR-SUBG \cite{DBLP:conf/icml/DasGNTZHJM22} & GCN & Target Query Subgraph, KNN Query Subgraph \\
        & SR \cite{DBLP:conf/acl/ZhangZY000C22} & GAT & Question subgraph \\
        & DELFT \cite{DBLP:conf/www/ZhaoXQB20} & GAT & question-Answer subgraph \\
        & Exaqt \cite{jia2021complex} & GCN & Answer-time fact subgraph \\
        & Graph2Seq \cite{DBLP:journals/tnn/ChenWZ24} & GCN & entity-relation graph \\

    \midrule

    \multirow{14}{*}{\makecell[l]{Recommendation\\System}} 
        & DSKReG \cite{DBLP:conf/cikm/WangLFSY21} & GAT & user-item graph, item-item graph \\
        & SARC \cite{DBLP:conf/ictai/ZhangCX19} & GCN & User-Item Graph, User-Entity Weighted Graph, Item-Entity KG Subgraph \\
        & KLGCN \cite{DBLP:journals/eswa/WangLZW22} & GCN+GAT & user-item graph, item KG \\
        & CGAT \cite{DBLP:journals/tkde/LiuYXMWZ23} & GAT & user-item graph, item KG \\
        & KGAT \cite{DBLP:conf/kdd/Wang00LC19} & GAT & user-item-entity graph \\
        & KGIN \cite{DBLP:conf/www/WangHWYL0C21} & GAT & user-item graph, item KG \\
        & HAGERec \cite{DBLP:journals/kbs/YangD20} & GAT & user-item-entity graph \\
        & KCAN \cite{DBLP:conf/cikm/TuCWZZ0021} & GAT & user-item graph, item KG \\
        & JNSKR \cite{DBLP:conf/sigir/ChenZMLM20} & GAT & user-item graph, item KG \\
        & KGCN \cite{DBLP:conf/www/WangZXLG19} & GCN & User-entity Graph \\
        & KGNN-LS \cite{DBLP:conf/kdd/WangZZLZLW19} & GCN & user-item graph, item KG \\
        & DIKGNN \cite{DBLP:conf/cikm/TuQWZLZWZZ23} & GAT & User-Fund graph, fund KG \\
        & SDK \cite{DBLP:conf/cikm/LiuXLWCY23} & HGNN+GAT & user-item graph, user hypergraph, item hypergraph \\
        & HyperTeNet \cite{DBLP:conf/icdm/VijaikumarHS21} & HGNN+GCN & User-item-List KNN graph, User-item-List hypergraph \\

    \midrule

    \multirow{6}{*}{\makecell[l]{Natural Language\\Process}} 
        & JointLK \cite{DBLP:conf/naacl/SunSQZ22} & GAT & question-Answer subgraph \\
        & ReasoningLM \cite{DBLP:conf/emnlp/JiangZZLW23} & GFormer & BFS Sequencing \\
        & JAKET \cite{DBLP:conf/aaai/Yu0Y022} & GAT & entity graph \\
        & GOProteinGNN \cite{DBLP:conf/cikm/KalifaSR25} & GCN & Protein-Gene Ontology Terms graph \\
        & ASGARD \cite{DBLP:conf/acl/HuangWW20} & GAT & DocGraph, SegGraph \\
        & CogNLG \cite{DBLP:journals/es/LaiYFCWWC24} & GCN & Source Entity-Extension Entity graph \\

    \midrule

    \multirow{10}{*}{\makecell[l]{Drug-Drug\\Interaction}} 
        & MKG-FENN \cite{DBLP:conf/aaai/WuSHC024} & GCN & \makecell[c]{Drug-Chemical Entity graph, Drug-Substructure graph, \\ Drug-Drug graph, Drug-Molecular Structure graph} \\
        & LaGAT \cite{DBLP:journals/bioinformatics/HongLJL22} & GAT & Drug-Drug graph \\
        & SumGNN \cite{DBLP:journals/bioinformatics/YuHZGSX21} & GAT & k-hop local subgraph \\
        & KGNN \cite{DBLP:conf/ijcai/LinQWMZ20} & GCN & Sampling Subgraph \\
        & MUFFIN \cite{chen2021muffin} & GCN & Drug-Drug graph \\
        & MSKG-DDI \cite{chen2024effective} & GAT & Molecular graph, drug KG \\
        & MK-GNN \cite{gao2023medical} & GCN & Drug-Drug graph \\
        & DeepLGF \cite{ren2022biomedical} & GCN & Drug-Receptor Graph \\
        & HetDDI \cite{li2023hetddi} & GCN & Drug Molecule graph \\
        & GCLR \cite{an2024disease} & GCN & Disease-Drug Graph, Disease-Disease Graph, Drug-Drug Graph \\

    \bottomrule
  \end{tabularx}
\end{table*}

\begin{table*}[htbp]
  \centering
  \caption{Statistics of datasets for Knowledge Extraction.}
  \label{tab:balanced-stats}
  \small
  \renewcommand{\arraystretch}{1.3} 
  
  \begin{tabular*}{\textwidth}{@{\extracolsep{\fill}} l l c c c p{6cm} @{}}
    \toprule
    \textbf{Task} & \textbf{Dataset} & \textbf{Train} & \textbf{Dev} & \textbf{Test} & \textbf{Source} \\ 
    \midrule

    \multirow{6}{*}{\textbf{NER}} 
        & Weibo        & 73.8 K  & 14.5 K  & 14.8 K  & Social media platform Weibo (\url{https://huggingface.co/datasets/Aunder/weibo_ner}) \\
        & Resume        & 124.1 K & 13.9 K  & 15.1 K  & Resumes collected from Sina Finance (\url{https://github.com/singhsourabh/Resume-NER}) \\
        & OntoNotes     & 491.9 K & 200.5 K & 208.1 K &  Sourced from News domain (\url{https://github.com/4AI/LS-LLaMA/tree/main})\\
        & E-commerce   & 119.1 K & 14.9 K  & 14.7 K  & Products and brands by crawling (\url{https://huggingface.co/datasets/mach-12/ecommerce-ner-conll-2003-data}) \\
        & Twitter-2015 & 6,176   & 1,546   & 5,078   & Social media platform Twitter (\url{https://github.com/1429904852/R-GCN}) \\
        & Twitter-2017 & 6,049   & 1,324   & 1,351   & Social media platform Twitter (\url{https://github.com/1429904852/R-GCN}) \\
    
    \midrule

    \multirow{4}{*}{\textbf{RE}} 
        & TACRED       & 68,124  & 22,631  & 15,509  & TAC KBP Competition (\url{https://github.com/XueFuzhao/GDPNet}) \\
        & Semeval      & 7,200   & 800     & 2,717   & Derived from international semantic evaluations (\url{https://github.com/cbaziotis/datastories-semeval2017-task4}) \\
        & DocRED       & 3,053   & 1,000   & 1,000   & Based on Wikipedia (\url{https://github.com/nanguoshun/LSR}) \\
        & CDR          & 500     & 500     & 500     & Sourced from PubMed (\url{https://github.com/fenchri/edge-oriented-graph/tree/reproduceEMNLP}) \\

    \midrule

    \multirow{4}{*}{\makecell[l]{\textbf{Joint}\\\textbf{Extract}}} 
        & NYT          & 56,196  & 5,000   & 5,000   & From The New York Times (\url{https://github.com/Coopercoppers/PFN}) \\
        & WebNLG       & 5,019   & 500     & 703     & Sourced from DBPedia (\url{https://github.com/Coopercoppers/PFN}) \\
        & SciERC       & 1,861   & 275     & 551     & AI scientific paper abstracts (\url{https://github.com/Coopercoppers/PFN}) \\
        & ACE05          & 10,051   & 2,424 & 2,050     & Sourced from PubMed (\url{https://github.com/Coopercoppers/PFN}) \\

    \bottomrule
  \end{tabular*}
\end{table*}

\begin{table*}[htbp]
  \centering
  \caption{Statistics of datasets for Knowledge Fusion.}
  \label{tab:knowledge-fusion-stats}
  \small 
  \renewcommand{\arraystretch}{1.3} 
  \setlength{\tabcolsep}{0pt}

  \begin{tabular*}{\textwidth}{@{\extracolsep{\fill}} l l c c c p{6.5cm} @{}}
    \toprule
    \textbf{Task} & \textbf{Dataset} & \textbf{Train} & \textbf{Dev} & \textbf{Test} & \textbf{Source} \\ 
    \midrule

    \multirow{5}{*}{\textbf{\makecell[l]{Entity\\Alignment}}} 
        & DBP15K        & 4,500   & \textbackslash & 10,500  & Multi-lingual versions of DBpedia (ZH/JA/FR-EN) (\url{https://github.com/cambridgeltl/eva}) \\
        & DWY100K       & 30,000  & \textbackslash & 70,000  & DBpedia, Wikidata, and YAGO (\url{https://github.com/cambridgeltl/eva}) \\
        & FB15K-DB15K   & 2,341   & 1,170          & 8,200   & Freebase and DBpedia (\url{https://github.com/Bubble-bubble77/EIEA/tree/main}) \\
        & FB15K-YAGO15K & 2,476   & 1,238          & 8,665   & Freebase and YAGO (\url{https://github.com/Bubble-bubble77/EIEA/tree/main}) \\
        & DICEWS        & 74.8 K  & 8.9 K          & 8.9 K   & Integrated Crisis Early Warning System (\url{https://github.com/lcai2/STEA/tree/main})\\
    
    \midrule

    \multirow{5}{*}{\textbf{\makecell[l]{Entity\\Linking}}} 
        & AIDA-CoNLL    & 18.5 K  & 4.8 K          & 4.5 K   & Derived from CoNLL 2003 Reuters News (\url{https://github.com/nicola-decao/efficient-autoregressive-EL}) \\
        & MSNBC         & \textbackslash & \textbackslash & 739     & Selection of news from MSNBC (\url{https://github.com/lephong/mulrel-nel}) \\
        & AQUAINT       & \textbackslash & \textbackslash & 727     & Subset of AQUAINT news corpus (\url{https://huggingface.co/datasets/naist-nlp/aquaint})\\
        & ACE2004       & \textbackslash & \textbackslash & 306     & Subset of ACE 2004 coreference data (\url{https://github.com/kpich/ace2004parse}) \\
        & BC5CDR        & 500     & 500            & 500     & PubMed abstracts (Chemical/Disease relations) (\url{https://github.com/Stubborn-z/MMR-FEK}) \\

    \bottomrule
  \end{tabular*}
\end{table*}

\begin{table*}[htbp]
\centering
\footnotesize 
\caption{Statistics of the datasets used in KGE and KR.}
\label{tab:dataset_stats}
\begin{tabularx}{\textwidth}{@{} l c c c c c c c X @{}}
\toprule
\textbf{Dataset} & \textbf{Entities} & \textbf{Relations} & \textbf{Hyperedges} & \textbf{Timestamps} & \textbf{Train} & \textbf{Dev} & \textbf{Test} & \textbf{Source} \\ \midrule
FB15k-237  & 14,541  & 237 & \textbackslash & \textbackslash & 272,115   & 17,535  & 20,466  & Freebase subset (inverse relations removed) (\url{https://github.com/zjukg/EARL/tree/master}) \\
WN18RR     & 40,943  & 11  & \textbackslash & \textbackslash & 86,835    & 3,034   & 3,134   & WordNet subset (inverse relations removed) (\url{https://github.com/zjukg/EARL/tree/master}) \\
YAGO3-10   & 123,182 & 37  & \textbackslash & \textbackslash & 1,079,040 & 5,000   & 5,000   & Entities with at least 10 relations in YAGO3 (\url{https://github.com/zjukg/EARL/tree/master}) \\
NELL-995   & 75,492  & 200 & \textbackslash & \textbackslash & 154,213   & 5,000   & 5,000   & Released from the NELL system iterations (\url{https://github.com/bdi-lab/InGram})\\ \midrule 
ICEWS14    & 6,869   & 230 & \textbackslash & 365            & 72,826    & 8,941   & 8,963   & Event-based temporal KG from Integrated Crisis Early Warning System (\url{https://github.com/JaySaligia/PPT}) \\
ICEWS18    & 23,033  & 256 & \textbackslash & 304            & 373,018   & 45,995  & 49,545  & Event-based temporal KG from Integrated Crisis Early Warning System (\url{https://github.com/JaySaligia/PPT}) \\
ICEWS05-15 & 10,094  & 251 & \textbackslash & 4,017          & 368,962   & 46,275  & 46,092  & Long-term temporal subset of ICEWS (\url{https://github.com/JaySaligia/PPT}) \\
GDELT      & 7,691   & 240 & \textbackslash & 2,751          & 1,734,399 & 238,765 & 305,241 & Global Database of Events, Language, and Tone news datasets (\url{https://github.com/INK-USC/RE-Net}) \\
YAGO       & 10,623  & 10  & \textbackslash & 189            & 161,540   & 19,523  & 20,226  & Temporal knowledge graph facts extracted from YAGO3 (\url{https://github.com/INK-USC/RE-Net})\\ \midrule
DBLP       & 43,413  & \textbackslash & 22,535 & \textbackslash & 18,028    & 2,253   & 2,254   & Co-authorship network (\url{https://github.com/26hzhang/DBLPParser}) \\
Cora       & 2,708   & \textbackslash & 1,072  & \textbackslash & 857       & 107     & 108     & Citation network (\url{https://github.com/DM2-ND/CFLP}) \\
Pubmed     & 19,717  & \textbackslash & 7,963  & \textbackslash & 6,370     & 796     & 797     & Medical publication network (\url{https://github.com/DM2-ND/CFLP})\\
Citeseer   & 3,312   & \textbackslash & 1,079  & \textbackslash & 863       & 107     & 109     & Scientific literature network with hypergraph structure (\url{https://github.com/DM2-ND/CFLP}) \\
JF17K      & 29,177  & 327            & 102,648 & \textbackslash & 76,379    & 1,701   & 24,568  & A subset of Freebase processed for $n$-ary relational learning (\url{https://github.com/UM-Data-Intelligence-Lab/NYLON_code}) \\ \bottomrule
\end{tabularx}
\end{table*}

\end{document}